\documentclass{article}
\usepackage{frExamplee}
\usepackage{graphicx}
\usepackage{apalike}
\usepackage{setspace}

\usepackage[compact]{titlesec}
\titlespacing*{\section}{0pt}{0.3\baselineskip}{0.3\baselineskip}
\titlespacing*{\subsection}{0pt}{0.2\baselineskip}{0.2\baselineskip}
\titlespacing*{\subsubsection}{0pt}{0.1\baselineskip}{0.1\baselineskip}
\usepackage[table,xcdraw]{xcolor}
\usepackage{threeparttable}
\usepackage{wrapfig}
\usepackage{lipsum}
\usepackage{multirow}
\usepackage{graphicx}

\usepackage{amsmath}
\usepackage{enumitem}
\usepackage[margin=0.5cm]{caption}
\usepackage[ruled,lined,boxed,commentsnumbered] {algorithm2e}
\usepackage{url}
\title{Bumblebee: A Path Towards Fully Autonomous Robotic Vine Pruning}
\author{
Abhisesh Silwal\thanks{ } \\
Robotics Institute\\
Carnegie Mellon University\\
Pittsburgh, PA, 15217 \\
\texttt{asilwal@andrew.cmu.edu} \\
\And
Francisco Yandun \\
Robotics Institute\\
Carnegie Mellon University\\
Pittsburgh, PA, 15217
\And
Anjana Nellithimaru \\
Robotics Institute\\
Carnegie Mellon University\\
Pittsburgh, PA, 15217
\AND
Terry Bates \\
CLEREL\\
Cornell University\\
Portland, New York, 14769
\And
George Kantor \\
Robotics Institute\\
Carnegie Mellon University\\
Pittsburgh, PA, 15217
}

\begin{document}

\maketitle

\begin{abstract}
Dormant season grapevine pruning requires skilled seasonal workers during the winter season which are becoming less available. As workers hasten to prune more vines in less time amid to the short-term seasonal hiring culture and low wages, vines are often pruned inconsistently leading to imbalanced grapevines. In addition to this, currently existing mechanical methods cannot selectively prune grapevines and manual follow-up operations are often required that further increase production cost. In this paper, we present the design and field evaluation of a rugged, and fully autonomous robot for end-to-end pruning of dormant season grapevines. The proposed design incorporates novel camera systems, a kinematically redundant manipulator, a ground robot, and novel algorithms in the perception system. The presented research prototype robot system was able to spur prune a row of vines from both sides completely in 213 sec/vine with a total pruning accuracy of 87\%. Initial field tests of the autonomous system in a commercial vineyard have shown significant variability reduction in dormant season pruning when compared to mechanical pre-pruning trials. The design approach, system components, lessons learned, future enhancements as well as a brief economic analysis are described in the manuscript. 

\end{abstract}

\section{Introduction}
\label{sec::intro}
Pruning is a primary tool used by grape growers to manipulate vine size and shape which helps to regulate crop-load and maintain vine balance. Dormant season grapevine pruning involves removal of plant tissues in the form of spurs and excess one-year-old canes from the previous year’s growth. It is a highly labor-intensive task that requires skilled workers during winter season, which are becoming less available. As labor workers are paid per vine to prune, the short-term seasonal hiring culture often leads to workers rushing to prune more vines in less time. This leads to inconsistent pruning of vines that often results in over/under cropping and could take several years of careful mitigation to recover and remain profitable \cite{bates2009}. Different pruning strategies have been extensively studied in various grape growing regions and grape varieties to achieve sustainable vine vegetative and reproductive growth, often referred to as vine or vineyard balance \cite{howell2001sustainable}. Some of the contemporary vineyard mechanization systems in vineyards during the dormant season include (in sequence) mechanical pre-pruning, manual pruning follow-up, and mechanical shoot or fruit thinning to maintain vineyard balance \cite{bates2014mechanical}.

Grapes are the leading fruits based on the production volume in the United States \cite{usda2019}. Its grape industry mainly consists of wine, table, juice, and raisin varieties that combined produced around 7.4 million tons of produce in 2017 \cite{usda2019}, \cite{ers2016} and is currently valued at the U.S. \$6.6 billion. Despite its impressive growth in the last decade, grape industry continues to rely on hand labor for many operations. Among the most labor-intensive and costly tasks in grape production are harvesting, pruning, cluster thinning and equipment operation. Pruning is often labeled as one of the top three costly tasks that could take up a quarter of labor costs in the fruit production cycle \cite{johnson2016}. According to the University of California Cost and Return Studies in 2017 \cite{Fidelibus2018,Alston2018}, table grape growers can annually incur operating costs up to \$18,000 per acre to generate income of about \$30,000 per acre \cite{Fidelibus2018}. These accounts to approximately 45 percent of costs just for labor. Future projection of the labor issue is expected to become more critical both in terms of uncertainty in the availability and increasing costs \cite{fennimore_doohan_2008,calvin2010}. These concerns about labor supply have promoted renewed focus and enhanced interest in mechanization and the use of advanced technologies to secure long-term sustainability of grape and fresh fruit industry, in general.

To reduce labor cost, vineyard mechanization research has played an important role in the grape processing industry in the U.S. The invention and adoption of the mechanical grape harvester in the early 1970s eliminated hand harvest as a labor issue in the grape juice industry. Research and development of mechanical pruning has continued since the mid-1970’s \cite{morris2007} and it alone has further reduced labor costs. However, the lack of specificity in retained nodes causes the vines to be over-cropped (out of balance) with poor fruit quality \cite{bates2008}, \cite{bates2009}. This lack of selective pruning capability only provides a partial solution as additional follow operations are often required to complete the task that further increases production cost. Grapevines are perennial plants with indeterminant growth habits, so canopy structures are highly vigorous, and the entanglement of canes quickly lead to canopies that are too complex to analyze even for trained human eyes, let alone for computer vision algorithms. Thus, a robotic pruner as a follow-up operation after mechanical pre-pruning could be a pragmatic solution. This work presents a systematic approach to integrate robotic systems to fully automate hand follow-up operations after mechanically pre-pruning. Further, the profit margins for commercial vine production in general are low, the quantity and quality of manual labor is declining while the cost for fuel and fertilizers are ever-increasing \cite{uzes2016factors}. The development of automated robotic pruning as the mechanical pruning follow-up operation would further reduce labor costs and increase specificity in retained node quantity and quality. The growing region considered is located in western New York, one of the largest juice grape producers in the U.S.

Our long-term goal is to develop a commercially viable and fully autonomous pruning system to reduce dependency on seasonal semi-skilled workers while improving productivity. The overall objective is to investigate robotic technology to significantly improve and stabilize the balance between vegetative and reproductive growth that would yield better fruit quality and predictable crop load. Our approach deviates significantly from the established paradigm in agricultural robotics in two major ways. Firstly, it is recognized that a grapevine training system that facilitates robotic technology in vineyards is key to successful implementation of autonomous and selective pruning of vines. Thus, the commercial vineyard in our study was specifically designed and is constantly modified to facilitate automation. Second, the design of the proposed robot is multi-functional with capability to perform other tasks such as autonomous multi-sensor data collection throughout the growing seasons while remaining compatible with different varieties and canopy architectures of vines that adds more novelty to existing systems and potential for commercialization. 

The remainder of this paper is organized as follows: Sections \ref{sec::review} discusses prior work in this field and how the outcomes of the previous research motivated some design selections. In Section \ref{sec::mods}, we describe the work environment modifications and basic viticultural terminologies for context. A key requirement for accurate perception for vine modeling and manipulation in complex environments in the outdoors was a robust illumination invariant camera system. The inclusion of such camera system to measure thin vine structures in this systems integration work is based on \cite{silwal2021robust}. Similarly, the 3D computer vision pipelines to generate and process vine models are based on shortcoming of 2D methods previously reported work by \cite{botterill2017robot}. The camera design consideration along with navigation and manipulation pipelines for robotic pruning are detailed under Section \ref{sec::methods}. Section \ref{sec::results} reports the results and lessons learned from field-testing of the pruner in a commercial vineyard. Finally, some concluding thoughts and discussion on further improving the robustness of the current design are in Section \ref{sec::discuss}.

\section{Relevant literature}
\label{sec::review}
In the past several decades, research on the development and use of robotic systems for various agricultural tasks have been widely studied by the scientific community. Automated solutions for sowing seeding, monitoring or pest-detection are widely documented in the literature \cite{Gollakota2011,Diago2015,Ebrahimi2017,li2009}. These are complex systems designed to work in unstructured environments and changing lighting conditions \cite{bac2014,gongal2015}. One of the most targeted applications of robotics in agriculture is harvesting of fruits and vegetables. In a recent work, \cite{bac2014} reviewed 36 different robotic projects completed between the years 1985 and 2012. All reviewed projects were developed for fruit or vegetable harvesting. Historically, the limiting factor in perception has been to robustly detecting fruit under occlusion and uncontrolled natural illumination \cite{dean2011,li2011} while removing fruits without damaging and achieving picking speed comparable to human picker has been the major bottleneck in the manipulation side \cite{Botterill2017}. Despite the obvious advantage of automated pruning and the underlying commercial benefit, automated pruning on the other hand has not received much attention when compared to harvesting. The lack of research interest and progress could be attributed to the complexity of the task itself. 

For harvesting applications, the target fruits are generally easy to reach, and simple point-to-point paths are enough without the need for collision avoidance \cite{Botterill2017}. Pruning, on the other hand, presents significant challenges as the system not only has to detect the canopy structure but also measure topological parameters such as the location and orientation of the cutting points in the branches \cite{he2018,tabb2017}. As vine structures become more vigorous, the entanglement of multiple canes could easily become too complex to solve. In the past, very limited attempts have been made to design and evaluate a full scale robotic system for pruning vines. A robot system to spur prune grape vine was designed by Vision Robotics Corporation in 2015 \cite{VisionRobotics2020}. This commercial prototype used a stereo camera to identify and localize cut-point in canes and an industrial robot arm to prune highly manicured vine structures. However, the performance characteristics, design details on perception and manipulation are unknown and publicly not shared. A recent full-scale vine pruning prototype consisting of a robot arm, multi-camera system and over-the-row supportive structure for controlled lighting was proposed by \cite{Botterill2017}. This system generated models of vines for collision free manipulation and autonomously pruned a row of a vineyard. 

Robotic systems designed to interact with plants such as pruning of dormant vines require robust perception capabilities for motion planning and manipulation in unstructured environments. Before such interaction happens in robotic pruning, locating the pruning points is a necessary step, which itself is a challenging problem given that vines lack of consistent structures in their natural form. To automate the process of pruning point detection in vines, \cite{corbett2012} presented an AI-based expert system. It was based on rules defined by a viticulturist and used 3D topological features of the tree such as length, curvature, angle,  etc. in deciding to whether to keep or prune the branches. Similarly, \cite{katyara2021reproducible} used a combination of mean predictive histogram of gradients and statistical pattern recognition with K-means algorithm to classify pruning locations. These resent efforts used some form of optimization-based approach to identify pruning locations in vines. In pruning, the answer to where to make cuts is dictated by the pruning rules set by viticulturist. However, regardless of pruning rule, the number of buds retained plays an important role as the new parts of the vine (both vegetative and reproductive) emerges from the retained buds. To our best knowledge, only our work physically detects and associates buds individually to each cane for the pruning decision not only to closely resemble manual pruning but also to prevent accidental over-pruning.

Furthermore, to identify pruning locations in complex vine structures accurately, additional semantic understanding of the scene is required. For example, the segmentation of the canes from vine structures and the precise measurement of important topological parameters such as bud distribution and cane lengths. Getting a detailed semantic map of plants in real time and consistently in the outdoors has always been a bottleneck in pruning and perception-based agricultural robotics, in general \cite{kazmi2014indoor,houle2010phenomics}. One of the major factors affecting consistency in having such capability is the changing outdoor lighting conditions that affects image quality. Historically, to limit effects from changing outdoor illumination, researchers have relied on external structures with controlled lighting. Such as by \cite{cane1,botterill2013finding}, \cite{VisionRobotics2020} where they addressed the background and illumination challenges by employing a wheeled platform with controlled lighting that completely covered the vines during imaging. The large platform had to be pulled along the rows at low speeds, resulting in a complex and slow application for pruning. Similarly, \cite{Kicherer} presented two different approaches to avoid uncontrolled lighting conditions and the presence of the vines from another row in the background. First, manual segmentation of images using an artificial white background and secondly the use of a multi-camera system for depth reconstruction. A robotic system to measure tree traits by 3D reconstruction of a fruit-tree in field settings was also presented by \cite{Tabb1}. They measured parameters such as branching structure, branch diameter, length and angle with low mean square error but required extensive computation time (more than 5 minutes per tree) making it not suitable for real-time in-field applications.

In a similar application, \cite{Tabb2} present a super-pixel based image segmentation method for semantic segmentation in field environments for tree reconstruction and apple flower detection. It also involved using a mobile background unit and capturing hundreds of images per tree.  Furthermore, image-based cane segmentation and applied Gibbs sampling was used to recover 2D structures of a dormant season grape plant from images by \cite{cane1}. They also presented a quantitative comparison of their method with previous work on 2D cane structure extraction \cite{botterill2013finding}. Although their method performed well in detecting cane segments, it suffers from low precision due to its inability to detect branching points and hence ending up with disjoint cane segments. Their system also relied on a customized background screen in the field to perform foreground-background segmentation. In a similar study, \cite{pruning_weight1} presented an image-based cane segmentation method to assess pruning weight in a vineyard. They overcame the background segmentation challenge in outdoor environment by using a white background to avoid the presence of the vines from another row in the scene and also by taking images at night without any background. Their research was more focused on background-cordon-trunk-cane segmentation for pruning weight assessment rather than pruning point identification. To achieve consistent image exposure in any lighting condition, \cite{pothen2016automated} used high resolution stereo sensor with flash to predict yield in vineyards from image-based counting for grapes. The use of flash imagery in this study generated images with uniform white balance and had minimal effects from natural illumination. Our design of the camera system in this paper is motivated by this work. In a follow-up study, we show that the consistency in images not only facilitate classical computer vision algorithm but also tend to reduce the amount of data required to train deep-learning networks \cite{silwal2021robust}.

Another aspect that makes a robotic system especially valuable in agricultural applications is its capability to navigate around the environment. Agricultural fields normally have off-road terrains where any vehicle has to drive in a safe, socially predictable, and in some cases energy-efficient manner. Challenges including noise in the sensors, loss of traction, space constraints, among others make this task specially complicated. Depending on the application and the type of crops, various strategies using perception, planning and control have been studied to develop autonomous or semi-autonomous systems to drive in these scenarios \cite{mousazadeh2013technical,bechar2016agricultural}. The perception subsystem normally uses data from cameras, laser range finders, inertial sensors (IMU) or GPS receivers to obtain information about the environment and localize the robot within a map \cite{chen2015real} or use SLAM algorithms \cite{abouzahir2018embedding}. With the continuous improvement in computation capabilities, machine learning approaches have become popular for this task, mainly using visual sensors \cite{chen2020survey}. Additionally, a diverse group of planners and controllers have been designed and used to guide and command robots to navigate in Ag settings \cite{papadakis2013terrain,ding2018model}. For example, in \cite{chen2020survey} a local planner was combined with a custom control law for an Ackerman vehicle driving in a hazelnut orchard. In this case, both the planner and the controller were designed to account for the kinematic constraints of the vehicle as well as the space restrictions that limited the maneuverability. Other than custom control laws, the stability, accuracy, and smoothness in the navigation that predictive approaches provide made them especially suitable for agricultural applications \cite{ding2018model}. Furthermore, when the characteristics of the terrain strongly constrain the vehicle movement, predictive traction control strategies have arisen as a suitable solution \cite{sunusi2020intelligent}. The mentioned perception, planning and control approaches have been successfully implemented mainly for supervision and sensing tasks \cite{fountas2020agricultural}. However, little work has been reported in the integration of an autonomous navigation system to work alongside specific complex agricultural activities such as harvesting and pruning. In fact, the design and evaluation of a methodology for an integrated autonomous system capable to perform these tasks remains as a gap in field applications.

In  summary, because of very complex requirement in perception and actuation, extremely limited work and success has been seen in robotic pruning of grapevines and pruning in general. The existing prototypes rely on external physical structure (over the row platform) for acquiring images in the outdoors. This makes the robot’s ability to turn, enter, and row following in agricultural terrain extremely challenging and less pragmatic. Most importantly, the rigid frame designs further limits compatibility to different varieties and canopy architecture that could potentially limit commercial adoption. We believe that our rugged and modular robot equipped with an illumination invariant camera system and novel approach to perception and manipulation will lead to a pragmatic and economical solution for automated pruning. 

\section{Field environment and workspace modifications}
\label{sec::mods}
\begin{figure} [h]
    \centering
    \includegraphics[height=1.98in]{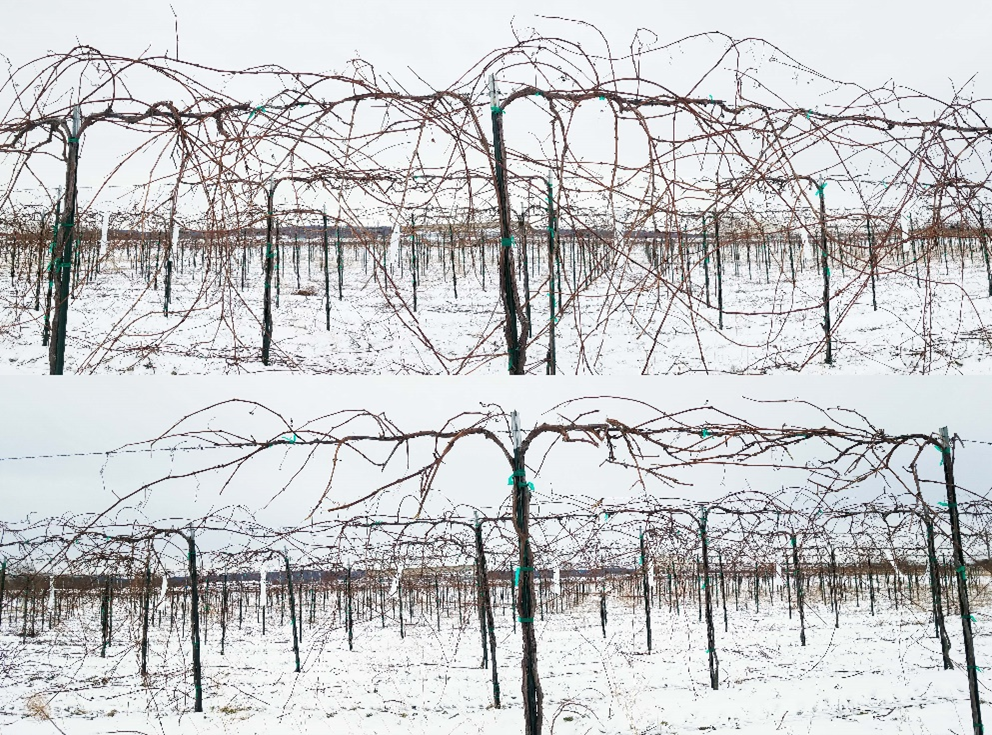}
    \hspace{.2in}
    \includegraphics[height=1.98in]{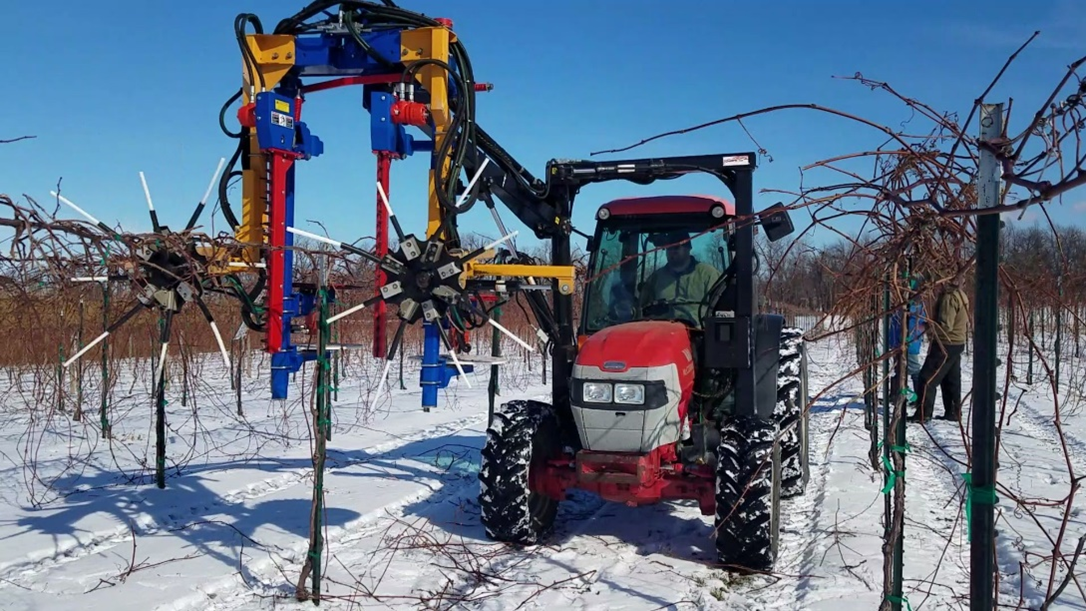}
    \caption{Dormant Concord grapevines before and after mechanical pre-pruning (left) with an OXBO V-Mech sprawl pre-pruner (right).}
    \label{fig:pre-pruing}
\end{figure}

The vineyard used for this study was located at the Cornell Lake Erie Research and Extension Laboratory in Portland, New York. Concord (Vitis labruscana, Baily) grapevines were own-rooted and planted in Chenango gravel-loam soil in 2012 at 2.6 m row by 2.4 m vine spacing and trained in bi-lateral cordon architecture with an average cordon height of 1.8 m. This variety of grape vine has indeterminant growth habits that results in canopy structures that are vigorous with high degree of cane entanglements, which creates a work environment where even manual pruning becomes a cumbersome task. A standard way in the industry eases labor intensity by mechanically pre-pruning the vines. The mechanical pre-pruners such as the VMech pre-pruning head comb grape canes up or down (white fingers Fig. \ref{fig:pre-pruing} right) into reciprocating cutter bars (red vertical Fig. \ref{fig:pre-pruing} right) that have an adjustable mechanism to retain longer or shorter canes. This mechanical pre-processing step, although a non-selective process, greatly minimized the complexity of the work environment. Following this industrial standard, in our experiments we mechanically pre-pruned the vines with an OXBO VMech 1210 Tool Arm and Sprawl pre-pruner (Vmech LLC, Fresno, CA). The mechanical pre-pruner was attached to a tractor and manually driven along the rows and was calibrated to remove canes greater than five nodes long. The result of mechanical pre-pruning and the pre-pruning machine are shown in Fig. \ref{fig:pre-pruing}.  
\begin{wrapfigure}{R}{0.35\textwidth}
\centering
\includegraphics[width=0.3\textwidth]{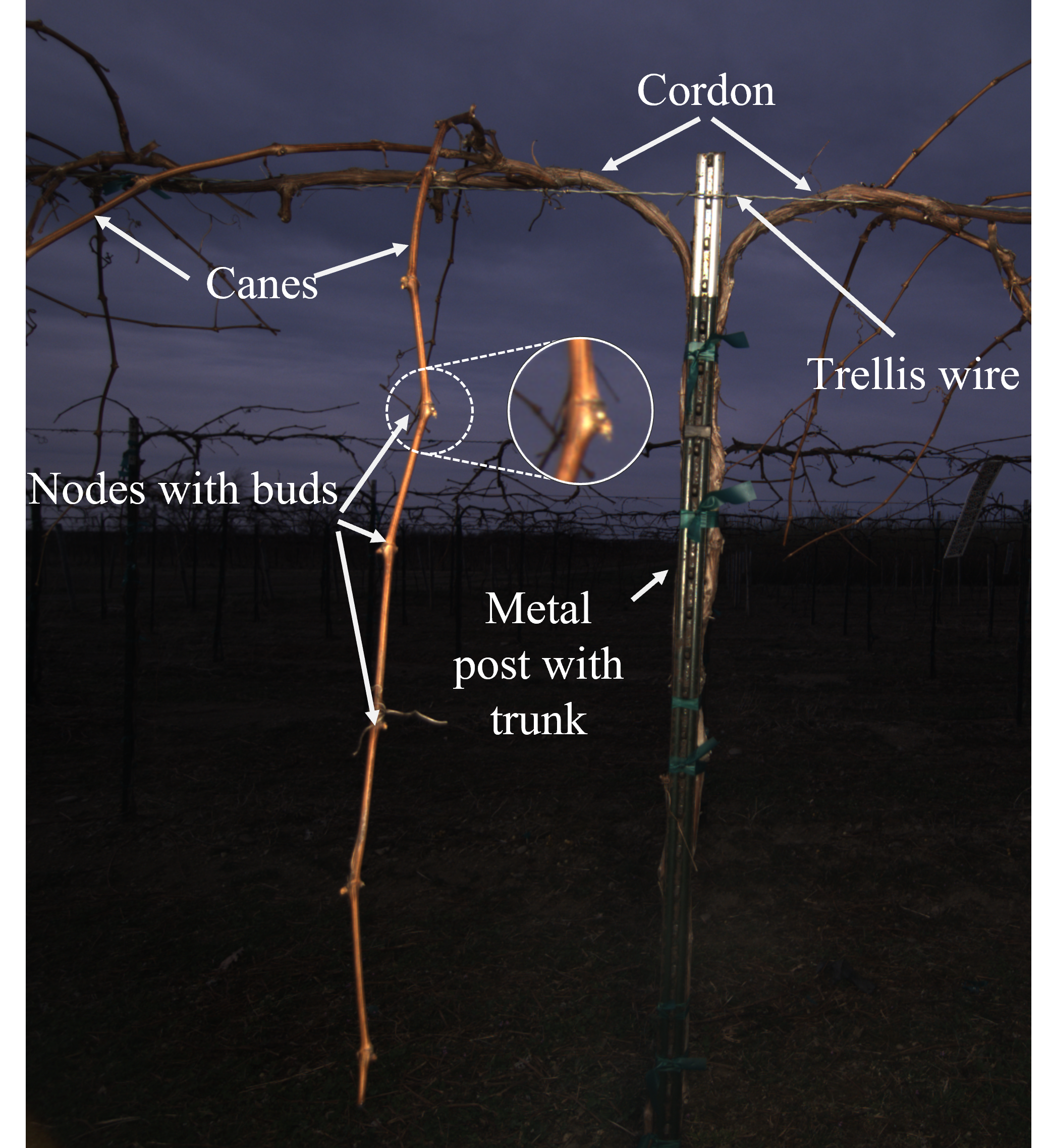}
\begin{minipage}[b]{0.28\textwidth}
\caption{Dormant grapevine canopy showing canes, cordon, trellis, buds, and nodes.}
\label{fig:vine_element}
\end{minipage}
\vspace{-100pt}
\end{wrapfigure}

Additionally, during our latest field trip (which was pushed towards the end of the pruning season because of the COVID-19 global pandemic), some of the vines that started to show vegetative growth were trimmed to retain its original dormant shape by removing the new shoots.

For context, the following brief definitions provide description of the vine canopy and commonly used terminologies in viticulture (see Fig \ref{fig:vine_element}).
\begin{itemize}[leftmargin=*]
\itemsep0em
\item \textbf{Bud}: \textit{ A bud is a growing point that develops in the leaf axils and often regarded as a compressed shoot.} 
\item \textbf{Shoot}: \textit{New green growth developing from a bud.}
\item \textbf{Cane}: \textit{A matured long, woody shoot after leaf fall.}
\item \textbf{Node}: \textit{The bulged part of a cane where buds are attached.}
\item \textbf{Cordon}:\textit{ The main lateral expansion of the trunk that supports shoots, canes, and fruits.}
\item \textbf{Pruning rule}: \textit{A set of rules that define a systematic way to remove older canes from grapevines.}
\end{itemize}

\section{Methods}
\label{sec::methods}
This section describes all components of Bumblebee. First, we describe the mechanical design of the robot that includes a prismatic base to increase the reachable workspace and then the design of the end-effector to prune vines. Secondly, we then detail the perception pipeline that describes the camera system, 3D reconstruction of vines from multiple views and novel 3D cane segmentation algorithm. The rest of the section  elaborates motion planning, navigation, and systems integration components.

\subsection{Mechanical design}
\label{sec::mechanical_Design}
\subsubsection{Manipulator}
\begin{figure} [h]
    \centering
    \includegraphics[height=2.0in]{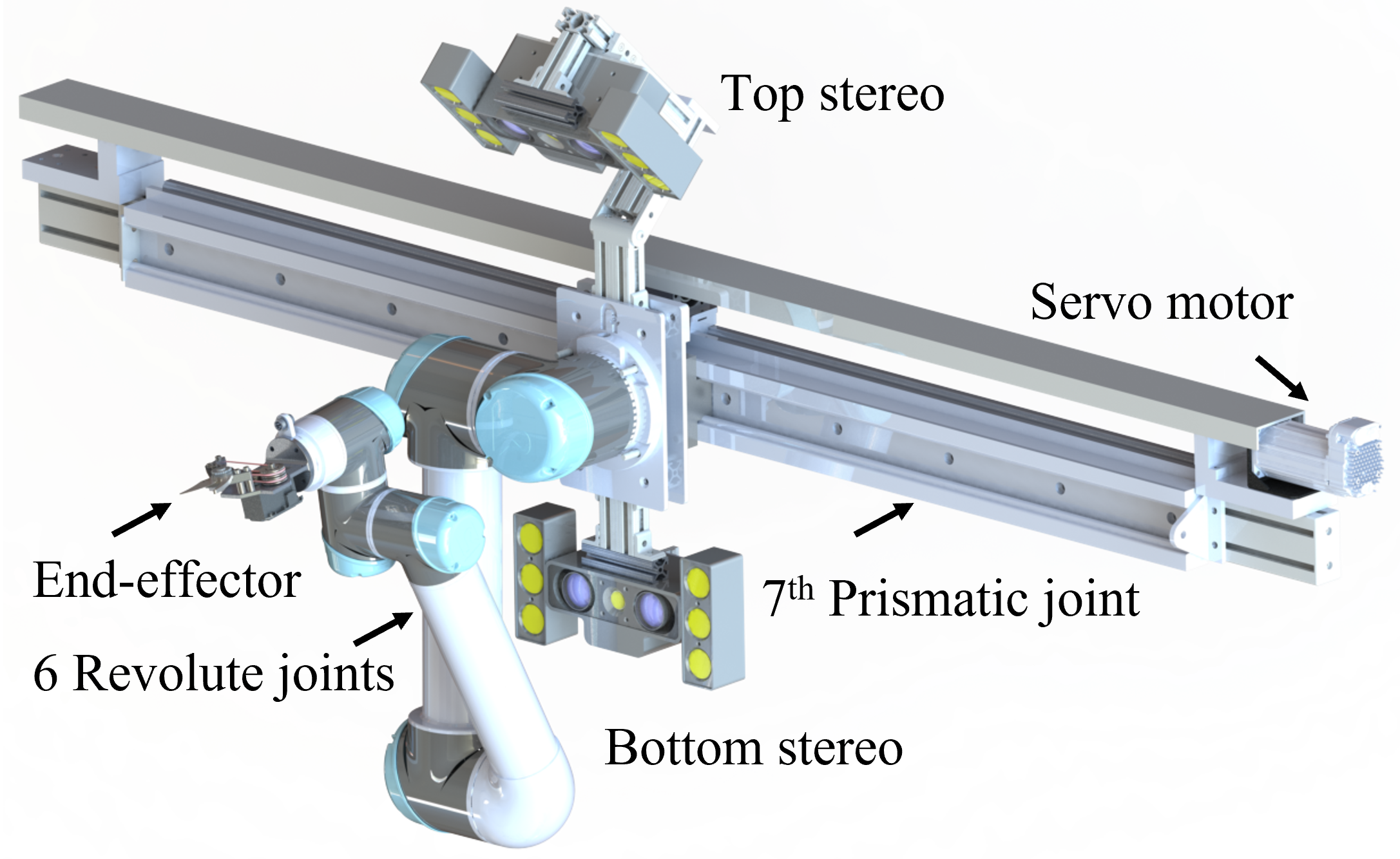}
    \hspace{.2in}
    \includegraphics[height=2.0in]{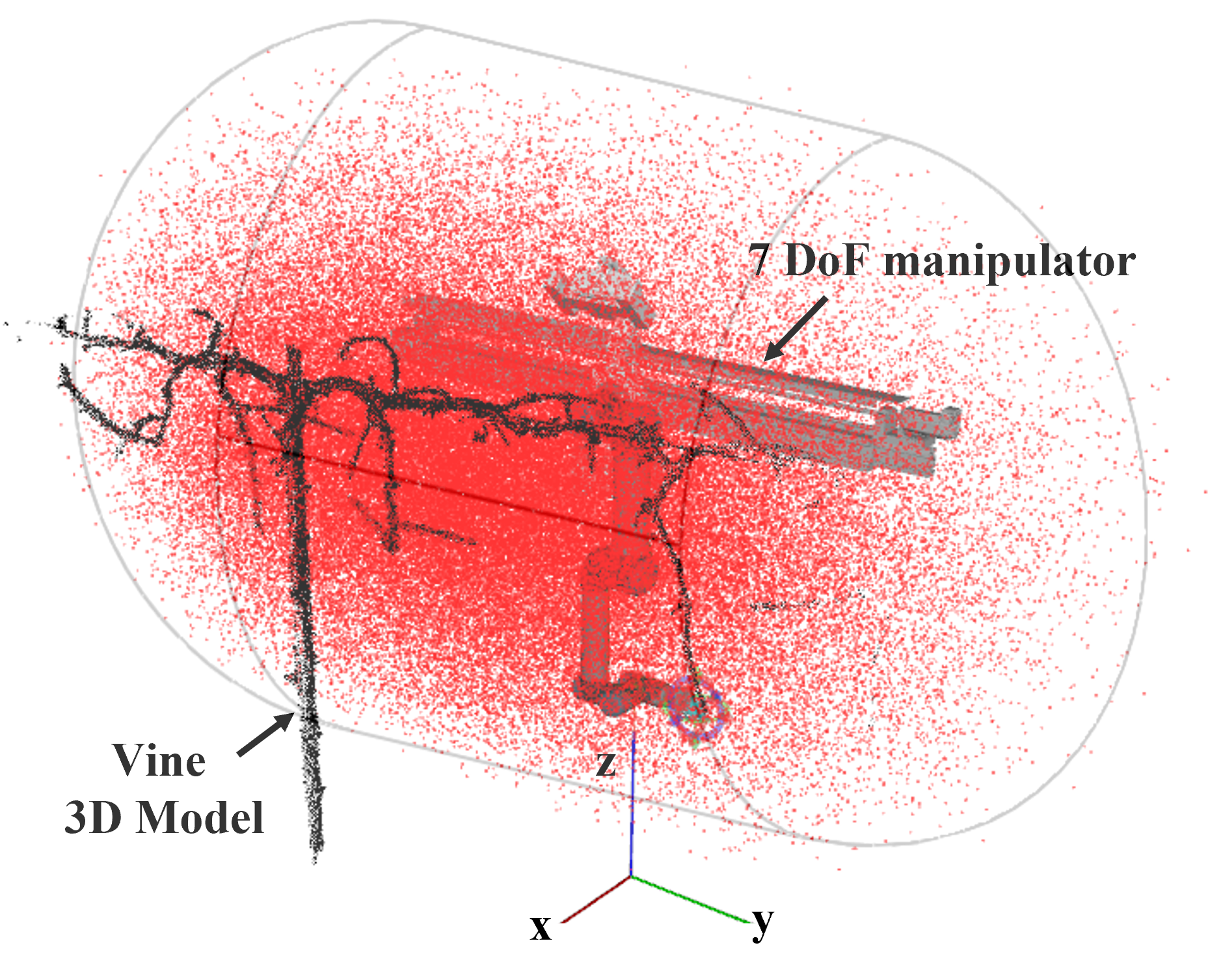}
    \caption{CAD rendering of the 7 DoF robot with its components (left). Reachable endpoint positions in the work envelope (right). A wire diagram is added to aid visualization.}
    \label{fig:robot_workspace}
\end{figure}

An unrestrained rigid body in 3D space has 6 Degrees of Freedom (DoF) described by the three translations and rotational angles about the three independent axes \cite{donald1984motion}. In theory, a robot arm with at least 6 DoF is required to achieve any pose in the workspace. In practice, this capability is severely limited by different factors such as singularities, self-collisions, collision models of the environment, etc. to name a few. However, in kinematically redundant mechanisms, the desired motion of the tool-end or the end-effector can be accomplished in an unlimited number of ways. In the design of the robotic manipulator for pruning dormant season vines, we extend the 6 DoF of an UR5 robot arm to 7 DoF by adding a prismatic joint to the base, as shown in Fig. \ref{fig:robot_workspace}, left. 

Since the DoF of the manipulator is greater than the Degrees of Constraints (DoC), our proposed design is kinematically redundant and offers several advantages. First, the kinematic redundancy physically allows the end-effector to achieve any combination of orientations required to reach pruning locations in a complex and unstructured work environment. Secondly, the motion of the robot arm is restricted by multiple constrains, such as joint limits and end-effector poses. These restrictions further narrow the convergence of the Inverse Kinematics (IK) and motion planning algorithms. Having redundancy greatly increases the odds of finding possible solutions and improves the convergence time and accuracy of these algorithms. And lastly, the span of the prismatic base drastically increases the reachable work envelope of the arm, as Fig. \ref{fig:robot_workspace}, right shows. This feature was particularly designed to provide the system with the ability to reach the entirety of the vine from a single stationary position without having to move the mobile base to new locations to work on the same vine. Thus, the possibilities of inducing errors in both manipulation and perception pipelines caused by the motion of the ground robot and the repetitive sensing of the same environment is highly reduced. The design of the redundant P6R open chain robot arm is shown in Fig. \ref{fig:robot_workspace} left. 

Although posing joint restrictions limits the full range of motion of each joint, constraints are important factors in motion planning. To control unnatural, unachievable, or unnecessary motions, several joint limits and constraints were set, as described in Table \ref{tab:robot_params}. A virtual wall (behind the linear base in Fig. \ref{fig:robot_workspace} left) imposed restrictions on the motion planner as target locations were always in front of the arm. This resulted in the “elongated hemispherical" shape of the work envelope for the 7-DoF arm. %Fig. 1c shows the difference between the work envelope of a static base 6-DoF arm versus the 7-DoF arm with a prismatic base.
\begin{table}[!h]
\caption{\label{tab:robot_params} Manipulator joint limits and other parameters.}
\centering
%\resizebox{\textwidth}{!}{%
\begin{tabular}{lcccc}
\rowcolor[HTML]{333333} 
{\color[HTML]{FFFFFF} Robot   Parameter} &
  \multicolumn{1}{l}{\cellcolor[HTML]{333333}{\color[HTML]{FFFFFF} Value}} &
  \multicolumn{1}{l}{\cellcolor[HTML]{333333}{\color[HTML]{FFFFFF} Joint   Name}} &
  \multicolumn{1}{l}{\cellcolor[HTML]{333333}{\color[HTML]{FFFFFF} Lower Limit}} &
  \multicolumn{1}{l}{\cellcolor[HTML]{333333}{\color[HTML]{FFFFFF} Upper Limit}} \\
Degrees of freedom     & 7                       & Base          & -0.675m   & 0.675m   \\
Max canopy depth reach & 0.9 m                   & Shoulder pan  & -2.43 rad & 2.01 rad \\
Prismatic  base length & 1.35 m                  & Shoulder tilt & -2.62 rad & 0.05 rad \\
Max joint velocity     & 0.15m/s \& 1.1rad/s*    & Elbow         & -2.62 rad & 0.05 rad \\
Max joint acceleration & 0.25m/s \&   0.25rad/s* & Writs 1       & -3.14 rad & 3.14 rad \\
Workspace volume       & 3.95 $m^3$                 & Wrist 2       & -3.14 rad & 3.14 rad \\
End-effector weight    & 0.5 kg (1.2 lbs.)       & Wrist 3       & -3.14 rad & 3.14 rad
\end{tabular}%
%}
    \begin{tablenotes}
      \item *Values used during the experiments.  
    \end{tablenotes}
\end{table}

\subsubsection{End-effector}
\label{sec::ee}
Pruning the grape vines require making precision cuts at a specific location in the canes. However, before such cuts could be made, it was important to understand the mechanical properties of the canes, especially the force required to cut dormant canes for the proper design of the pruning end-effector. Due to numerous environmental factors such as soil properties and access to nutrients and water, vines exhibit wide variation in the length and diameter of dormant canes \cite{bates2017}. In this experiment, we selected 30 samples of Concord canes with a wide range in sizes and age to quantify the force required for a successful cut. On average, the cane diameter of the collected samples varied from 5 mm to 11 mm. 
\begin{wrapfigure}{R}{0.45\textwidth}
\centering
\includegraphics[width=0.375\textwidth]{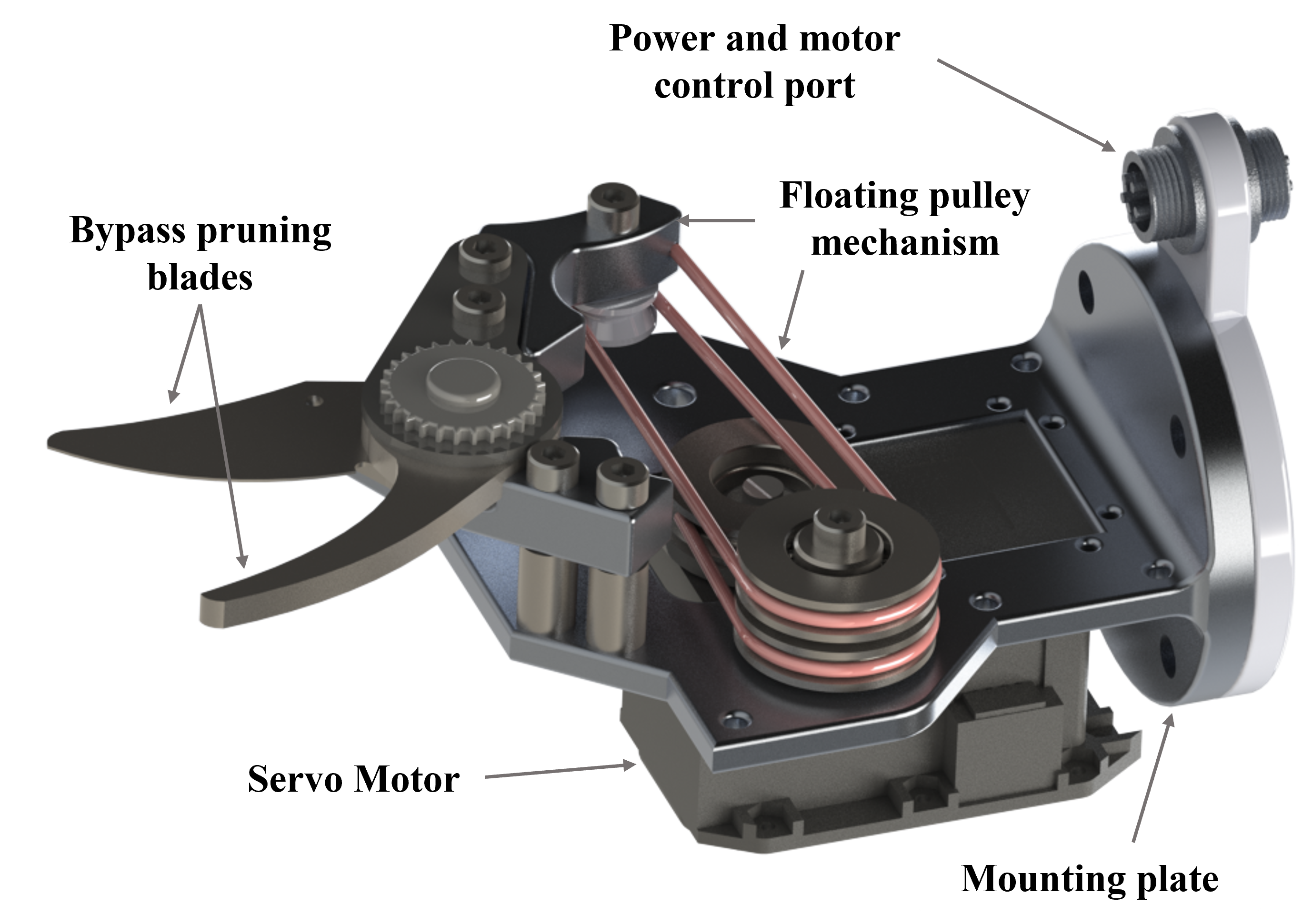}
\begin{minipage}[b]{0.4\textwidth}
\caption{A CAD model showing the design and components of the end-effector. A compression spring (not shown here) keeps tension on the pulley and has a mechanical advantage of 3.}
\label{fig:mechanical_design_end_effector}
\end{minipage}
\end{wrapfigure}

\begin{figure}[h]
    \centering
    \includegraphics[height=2.4in]{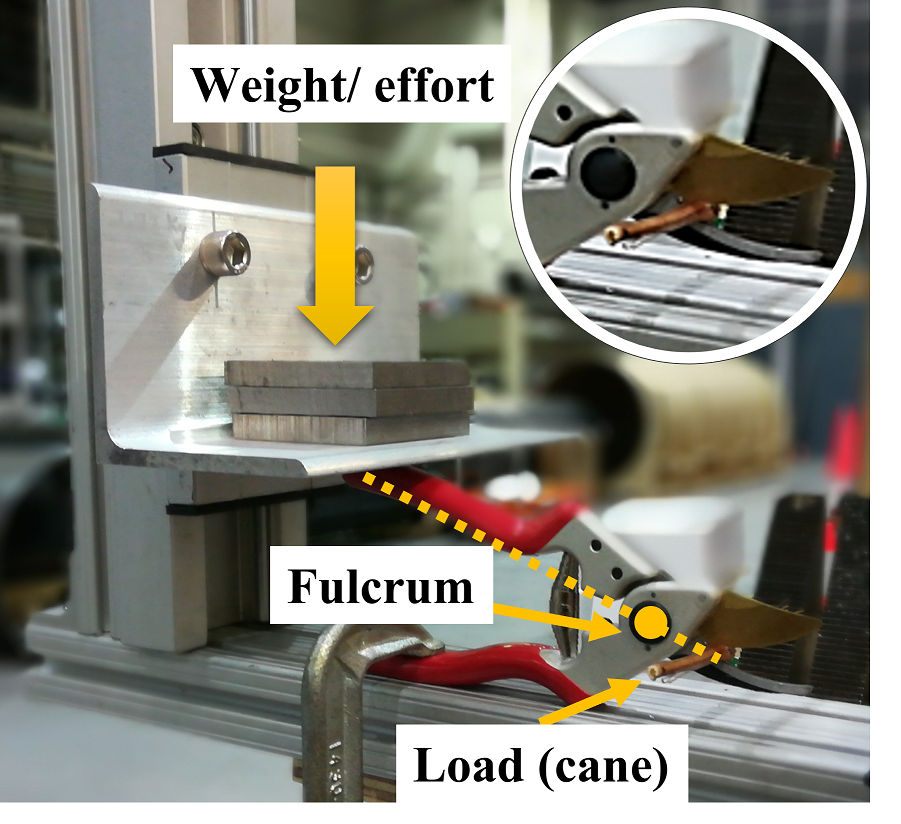}
    \hspace{.2in}
    \includegraphics[height=2.4in]{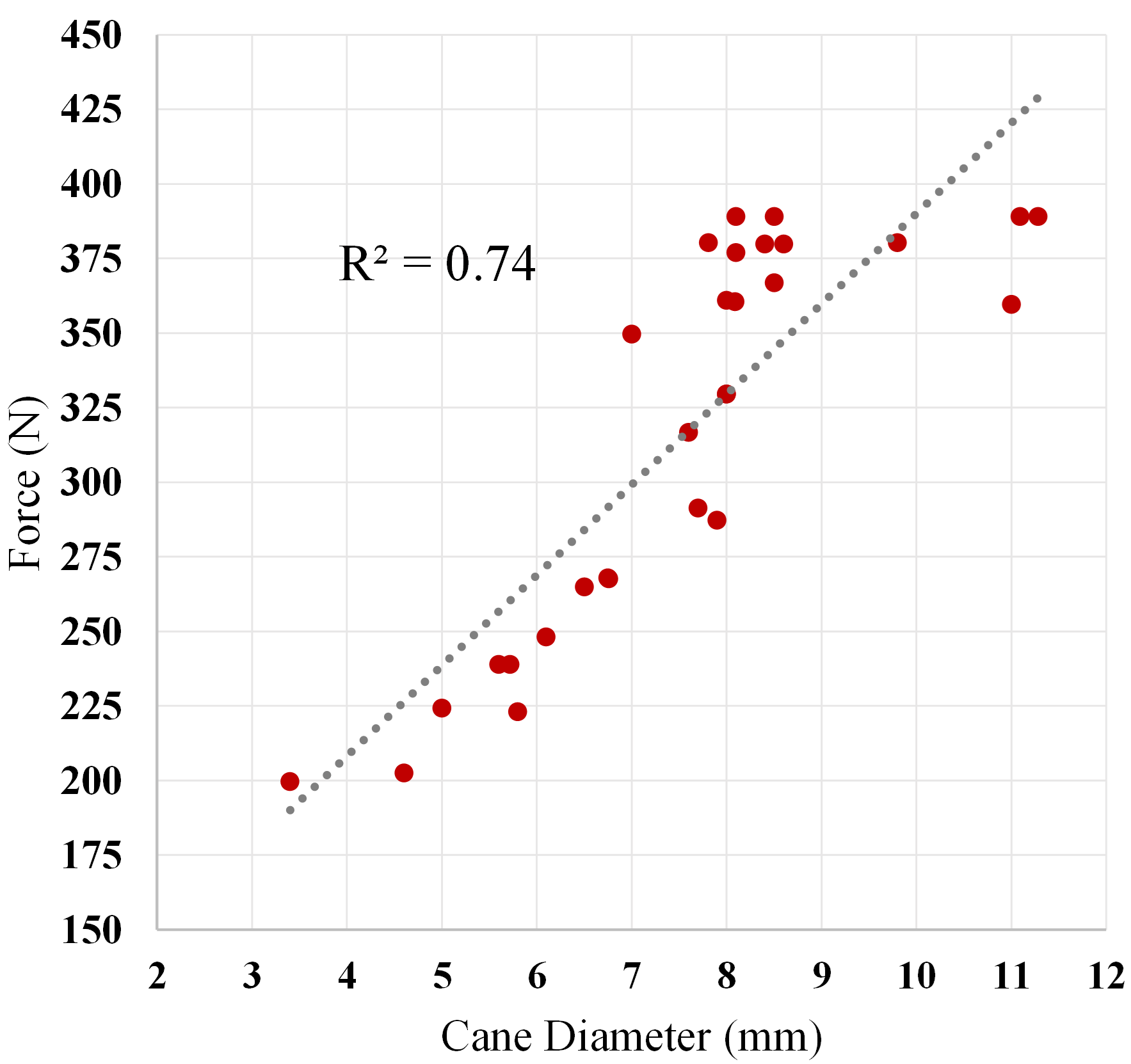}
    \caption{Experimental setup to measure force required to cut dormant canes (left). Correlation between cane diameter and cutting force (right).}
    \label{fig:cane_correlation}
\end{figure}

While the freshly cut live canes showed the presence of hard outer shell with soft internal tissues, dead samples exhibited relatively harder, shrunk, and dry internals and required higher force to cut. An experimental setup for this quantitative experiment is shown in Fig. \ref{fig:cane_correlation} (left). The mechanism of using hand-held shears to cut canes (like scissors) operates under the principles of the first-class lever, and it involves applying normal force on both handles at the same time. The experimental setup in Fig. \ref{fig:cane_correlation} (left) simplifies this requirement to just one normal force by fixing one of the handles to a rigid surface. An incremental load was then applied at the top of the movable handle to cut the cane samples, which were set at a fixed distance from the pivot and placed orthogonal to the cutting plane. Then, the total weight (load) along with the mechanical advantage of the lever at the abscission point provided the cutting force. This experiment was repeated on all sample canes. On average, 320 N force (at the abscission point) were required to cut a typical cane with 8 mm diameter. In Fig. \ref{fig:cane_correlation} (right), a fairly linear relationship ($R^2 = 0.74$) can be seen between cane diameter and the normal force required for cutting. 

A popular choice among professional pruners to prune grape vines is bypass pruning shears. This variety of pruning shears have blades that completely “bypass” each other for precise cuts and clean separations of the canes. Motivated by this pragmatic feature, the design of the end-effector includes a similar bypass mechanism (Fig. \ref{fig:mechanical_design_end_effector}). In this custom designed end-effector, one end of the scissors was fixed and bolted to the frame of the end-effector, whereas the other end was movable and actuated with a combination of a high torque (8 Nm) servo motor and floating pulleys. The floating pulley mechanism transferred power from the motor to the blades with a 200 lb (90.72 kg) fishing line. This simple machine system also added a mechanical advantage (M.A) of 3 and increased the overall factor of safety by nearly 8 folds. The combination of lightweight materials, simple machine, and high torque servo motor ensured a small (0.45 kg) yet powerful end-effector that fell well within the payload capacity (5 kg) of the robot arm. The small footprint and weight of the end-effector were also critical to ensure not only wide ranges of accelerations while executing motion trajectories but most importantly to get full horizontal extension of the manipulator into the canopy which could have been unattainable with heavier and larger end-effectors.
%\begin{figure} [h]
%    \centering
%    \includegraphics[height=2.75in]{images/cane_cutting_correlation.png}
%    \hspace{.2in}
%
%    \caption{Correlation between cane diameter and cutting force.}
%    \label{fig:cane_correlation}
%\end{figure}

\subsection{Perception}
\begin{figure} [h]
    \centering
    \includegraphics[height=1.5in]{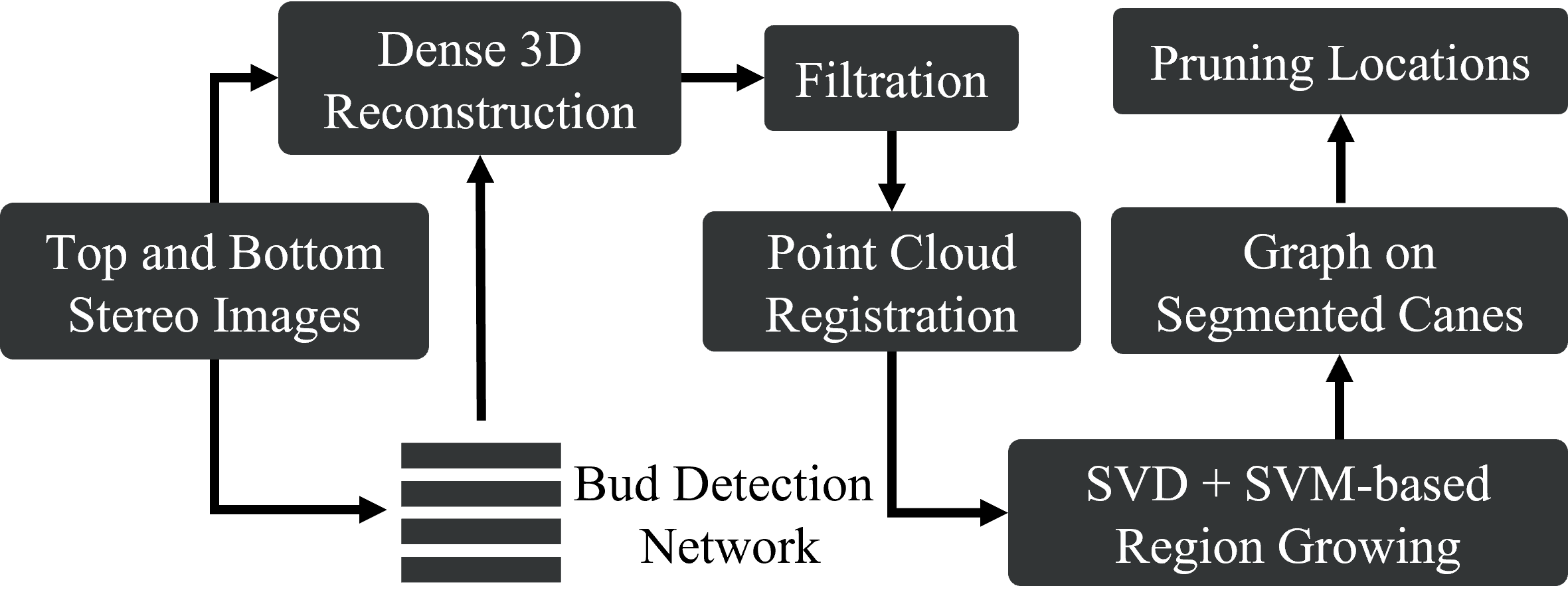}
    %\hspace{.2in}
    \caption{A block diagram with all major steps in the perception pipeline. Program flows from left to right.}
    \label{fig:perception_pipeline}
\end{figure}

The overall perception pipeline to perceive the vines and identifying the pruning locations is as shown in Fig. \ref{fig:perception_pipeline}. The major steps in the pipeline involve acquiring static images from fourteen  view points, point cloud registration using ICP algorithm, bud detection with Faster-RCNN, and cut-point detection with 3D region growing-based cane segmentation and a graph search algorithm.  

\subsubsection{Camera system} 
Performance of computer vision algorithms in the outdoor greatly depend on factors such as motion blur and changing illumination, to name a few. Among other, abrupt changes in the lighting condition can alter image quality that can lead to large data requirement in machine learning-based computer vision algorithms to compensate for variance in images \cite{silwal2021robust}. To minimize such effect, the camera system in our design uses active lighting that greatly minimizes the effects of varying environmental lighting while maintaining consistent image exposure. The detailed description, functionality and advantages of the active light camera system are described in our previous work \cite{silwal2021robust}. The camera system in this paper uses two of these cameras (Fig. \ref{fig:robot_system}) in a top-bottom stereo configuration and moved along the linear slide to image the vines from different view-points. The bottom stereo provided front views from a plane parallel to the vines, whereas the top stereo provided tilted views from higher elevation to include occluded parts deeper into the canopy not visible from just the front view. Altogether, the 3D information from multi-view geometry enabled us to generate accurate and complete 3D models of complex vine structures. The design of the dual stereo camera with the linear base is shown in Fig. \ref{fig:robot_workspace} (left) and a sample image-set of a vine with and without active lighting is shown below in Fig. \ref{fig:image_quality}. 

\subsubsection{Dense 3D reconstruction} 
\label{sec::camera}
One of the critical pieces of perceptual information required for autonomous pruning is precise and accurate 3D modeling of the vines. Once the 3D model is generated, the analysis of the plant geometry and topology can be automated, which also becomes essential for obstacle detection and ultimately in detecting of the pruning locations. To generate a dense 3D point cloud of the vines, the dual stereo camera imaged vines at seven precisely set positions along the liner slider. First, the individual point clouds from the top and bottom cameras at the seven different linear positions were registered separately ($PCL_{top}$ \& $PCL_{bottom}$) using Eqn. \ref{eqn:icp_linear}. To keep the point cloud registration time low, $PCL_{top}$ \& $PCL_{bottom}$ were registered in parallel. As show in Eqn.\ref{eqn:icp_top_bottom}, the final 3D model ($PCL_{final}$) of the vine was then constructed by registering $PCL_{top}$ and $PCL_{bottom}$ using the top-bottom camera extrinsic parameters (${T\ }_{top2bottom}$). Alternatively, each top and bottom point cloud pair could have been stitched in series to generate $PCL_{final}$. However, because of higher point cloud overlap and accurate initial transformation from the slider, this sequence of stitching individually registered $PCL_{top}$ \& $PCL_{bottom}$ resulted in better reconstructed model. %These processes are represented in Eqn. \ref{eqn:icp_linear} \& \ref{eqn:icp_top_bottom}.
\begin{equation}
\label{eqn:icp_linear}
    {PCL}_{\ top/bottom}\ =\ {PCL}_1\ +\ \sum_{i=2:n}{{PCL}_i\ \ast\ {T}_{i-1}} 
\end{equation}
  \begin{equation}
  \label{eqn:icp_top_bottom}
      {PCL}_{final}={PCL}_{bottom}+{PCL}_{top}\ast{T}_{top2bottom}
\end{equation}
Given two point clouds, fixed ($F$) and moving ($M$) where, $F=\left\lbrace f_1,f_2,\ldots,f_n\right\rbrace$; $f_i\in R^3$ and $ M=\{m_1,m_2\ldots,m_n\}$; $ m_i\in R^3$, the Iterative Closest Point (ICP) algorithm finds a rotation matrix $R$ and a translation vector $t$ such that error between transformed $M$ and $F$ point clouds is minimum. We take advantage of the precise initial transformation or correspondence between $F$ and $M$ provided by the linear sliding mechanism and calculate R and t in closed form. For this point cloud registration process, we experimented with the non-linear version of the point-to-plane ICP by \cite{fitzgibbon2003robust}. This variant of ICP provides more accurate estimation in cases where consecutive point clouds have different densities and exact correspondences are sparse \cite{fitzgibbon2003robust}. Although the point-to-plane ICP takes more time per iteration due to the added cost of point cloud normal computation, it usually converges in fewer iterations compared to classical ICPs. The point-to-plane ICP for our pipeline iteratively estimates $R$ and $t$ to minimize the distance between every point $m_i$ and the tangent plane at its corresponding point $f_i$. A tangent plane is represented by its unit normal $n_i$ computed around a small 3D neighborhood of 50 points in the vicinity of the point $f_i$. To reduce computation time, the size of each point cloud from the camera was reduced using voxel grid sampling and outliers were removed using statistical outlier filtering prior to ICP registration. The governing nonlinear equation is shown in Eqn. \ref{eqn:icp_nonlinear}
\begin{equation}
\label{eqn:icp_nonlinear}
    \min_{R,t} \sum_{i=1}^k ( | (R m_i+t-f_i)^T n_i| ^2 )
\end{equation}

\subsubsection{Bud detection} 
\label{sec:bud_detection}
A vine node usually consists of several buds (also known as compound bud) and has primary, secondary, and tertiary backups. During the growing season, if the primary bud is damaged for reasons such as external injury, frost bite or other environmental factors, vines sequentially release each remaining backup to replace the damaged/fallen buds. These buds are relatively small and are randomly positioned in the node, which makes it harder to detect in images and 3d models. Thus, despite the presence of multiple buds in a single node, detecting and counting buds as nodes is a reasonable approach. Moreover, this assumption is valid considering the fact that the secondary and tertiary buds generally bear insignificant amount of fruit compared to the primary \cite{budsBare}. In this work, the task of detecting buds is accomplished by detecting nodes. A node as described in Section \ref{sec::mods} is the bulged part of the cane that is more visible and has distinct features compared to the rest of the vine. From this point forward, we will use the terms bud and node interchangeably.

Detecting buds is a critical step in the pruning process of grape vines. Pruning rules such as cane and spur pruning which are popular in commercial settings involve retaining a certain number of buds per cane \cite{wsu2021}. Thus, accurate counting of buds is extremely important for autonomous pruning of dormant grape vines. To count buds, we leverage on the robustness of deep learning-based 2D object detection in the color images of the vines.  We used Faster-RCNN object detection network \cite{ren2015faster} to detect buds in one image from each stereo pair (top and bottom). For training, we used transfer learning  and initialized the network weights with the pre-trained \textit{imagenet} model before fine-tuning the network to our custom dataset. It consisted of 120 hand labeled images of buds collected prior to the field experiments. Although the number of images in the dataset seem small, the number of instances of buds per image were significantly larger. On average, 45 bud instances were present per image.

The detected buds in the 2D images (top and bottom) were then projected into the 3D space using the camera intrinsic parameters that produced sparse point clouds of the buds. This operation occurred in parallel to the point cloud registration process discussed in the above section and utilized the optimized ICP transformations for final registration. The combination of registered vine and bud point cloud (here after referred to as input point cloud, $PC_i$) completed the 3D vine modeling process. The bud-detection network datasets details and training parameters are listed in Table \ref{tab:bud_dataset} and \ref{tab:training_param} respectively. Figure \ref{fig:obstable_detection} show sample detections and 3D projection of the buds in the top camera image.
\begin{table}[!h]
\centering
\caption{\label{tab:bud_dataset} Dataset details for the bud detection network.}
\vspace{10pt}
\begin{tabular}{cccc}
\rowcolor[HTML]{C0C0C0} 
Network Name      & No. of images & Image size    & No. of buds \\
Faster-RCNN VGG16 & 120           & 2448   x 2048 & 5420    
\end{tabular}
\end{table}
\begin{table}[h]
\centering
\caption{\label{tab:training_param} Training parameters for the bud detection network.}
\vspace{10pt}
\begin{tabular}{ccccc}
\rowcolor[HTML]{E7E6E6} 
{\color[HTML]{000000} Training   samples} &
  {\color[HTML]{000000} Test images} &
  {\color[HTML]{000000} Learning rate} &
  {\color[HTML]{000000} Epochs} &
  {\color[HTML]{000000} Augmentation} \\
85 &
  35 &
  0.01 &
  300 &
  Horizontal   flip, scale, random crop
\end{tabular}
\end{table}
\subsubsection{Obstacle detection} 
\begin{figure} [h]
    \centering
    \includegraphics[height=2.250in]{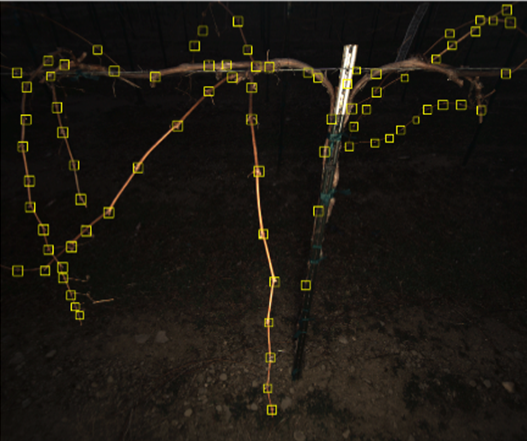}
    \hspace{.2in}
    \includegraphics[height=2.250in]{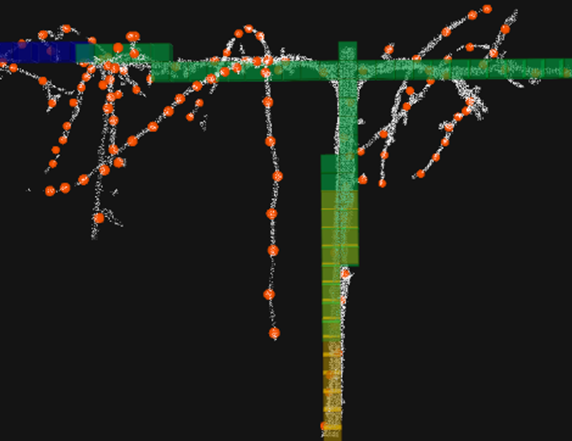}
    \caption{Bud detection with Faster-RCNN (left), Cordon and trellis wire as hard obstacles for motion planning (right)}
    \label{fig:obstable_detection}
\end{figure}

Using the dense 3D point cloud and non-linear point to plane ICP registration, we were able to consistently generate precise and clean point clouds of vines with buds that could be directly used for manipulation tasks. To avoid damage and reduce contact between the robot arm with the vine and its rigid support structure, it was necessary to define obstacles. In this work, only the central trunk with the metal post and horizontal cordons were taken as obstacles, as contact with theses rigid structures could potentially cause serious damage. However, as canes are relatively flexible and move when pushed, contacts between these soft objects and the robot arm were allowed to facilitate the motion planning (see Section \ref{sec::planning} for more details). Occupancy grid maps are popular choices to define occupied versus free spaces in the robot’s workspace. To define cordon and trellis wire as obstacles, a RANSAC algorithm \cite{derpanis2010overview} fitted two (vertical and horizontal) lines in the 3D model of the vines (Fig. \ref{fig:obstable_detection} right). As seen in Fig. \ref{fig:obstable_detection} (left), the new vine architecture has a vertical metal post to support the trunk and a horizontal trellis for the cordon to extend laterally. The existence of these features in the point cloud greatly benefited the RANSAC algorithm to precisely and consistently fit 3D lines in all vines used in our experiment. These fitted 3D lines were then the only elements taken as occupied space in the Octomap occupancy grid mapping algorithm \cite{wurm2010octomap} (Fig. \ref{fig:obstable_detection} right).

\subsubsection{Region growing for cane segmentation}
\label{sec::region_grow}
In general, the purpose of the region growing algorithms is to merge adjacent data points depending on a region membership criterion. In 3D point cloud space, this criterion could be smoothness constraints, resulting in points with similar smoothness profile clustered together. In our case, we use region growing algorithm for cane segmentation by clustering 3D points belonging to the canes. As our 3D structures of interest are thin canes, they did not have planar surfaces for normal computation to make use of a normal based smoothness constraint. Hence, we propose a novel region growing based segmentation method that utilizes local structural properties of the vines.  We used Singular Value Decomposition (SVD) \cite{stewart1993early}  on a small neighborhood of points of the vine's point cloud to understand the local structure of the vine in that small area. If there was only one dominant vector after SVD, that indicated the local neighborhood has a linear shape. If the number of dominant vectors was 2 or 3 that indicated the local shape is planar like and sphere like respectively.
Using this local shape information, we were able to segment out the linear portions of a cane from cane-cordon or cane-cane intersection regions which have non-linear 3D distribution in the local neighborhood.

\begin{figure} [h]
    \centering
    \includegraphics[height=1.8in]{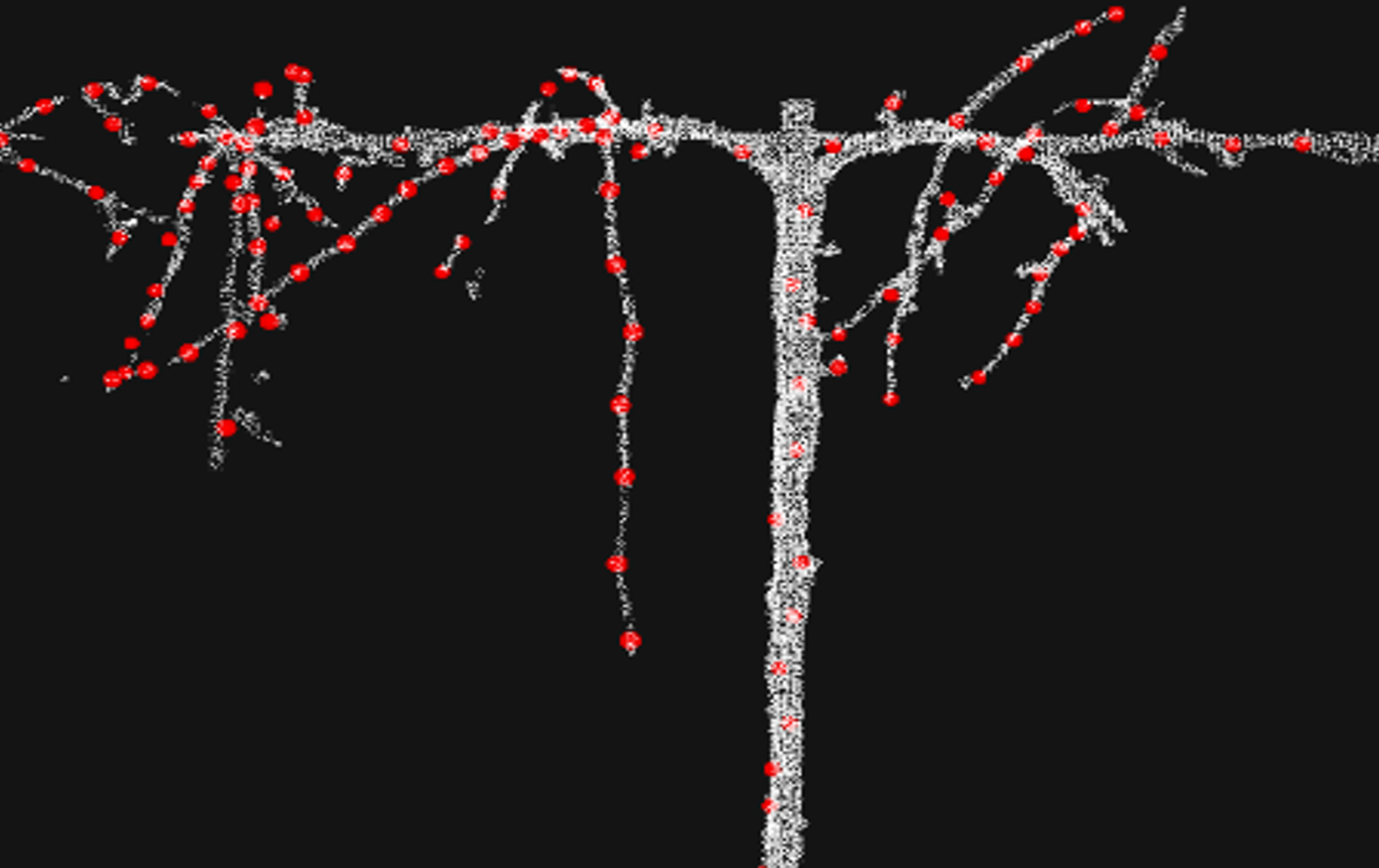}
    \hspace{.2in}
    \includegraphics[height=1.8in]{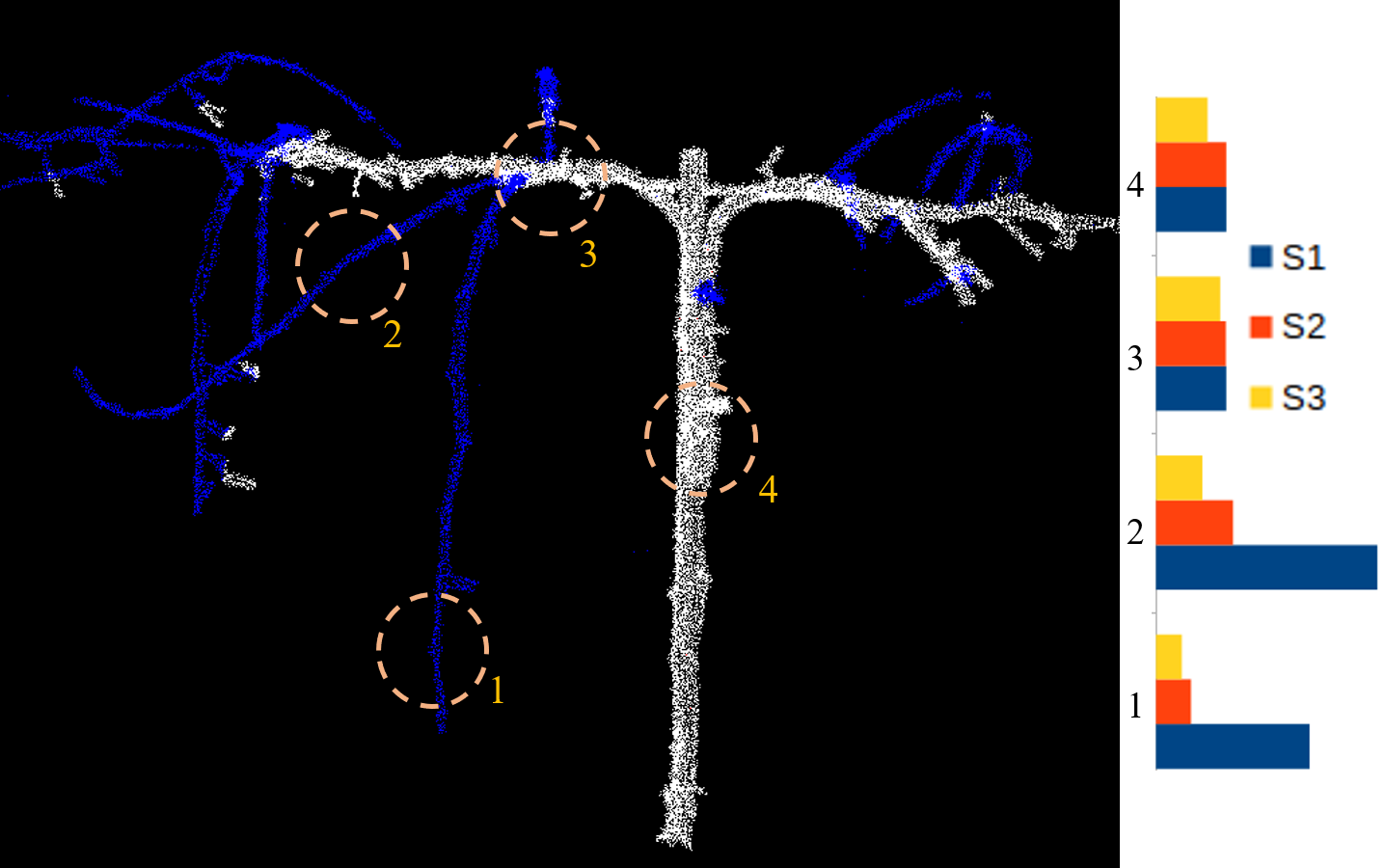}
    \caption{Vine 3D model with buds as seeding location for segmentation (left), segmented canes (center), and sample singular values after SVD (right).}
    \label{fig:segmentation}
\end{figure}

We started by randomly selecting a 3D projected bud location as a seed point. A set of points around the seed location were then extracted using a radial neighborhood search operation. This search radius (a hyperparameter) was pre-determined empirically based on the average distance between two consecutive buds in canes. Then, SVD calculated the singular vectors and values from the set of neighborhood points. For the extracted set of input points $N = (Pi) \in R_{mx3}$, the output singular values (non-zero diagonal elements) were $M = (Si) \in R_{1x3}$. The normalized singular values were then passed through a Support Vector Machine (SVM) that learned to classify patterns in the singular values (Fig. \ref{fig:segmentation}) as cane or intersection region. Based on the inference from the SVM, the set of extracted points were then labeled as a part of cane or intersection region between cane and cordon. This operation was iterated for all bud locations, and the steps are listed in the algorithm box \ref{alg:region_growing}. The training data for SVM involved 120 manually picked samples of cane and intersection regions. About 70\% (84 samples) were used for training and the remaining 30\% for test and validation. 

\SetKwComment{Comment}{/* }{ */}

\begin{algorithm}[hbt!]
%\LinesNumbered
\caption{Point cloud region growing}\label{alg:region_growing}
%    \SetAlgoLined
\KwData{Point cloud ${P}$, Directions={positive, negative}}
\KwResult{Point cloud ${R}$}
{$S \longleftarrow$ Bud Detections\;
$A \longleftarrow\{1,...,|P|\}$ - Cordon points\; 
\While{$S \neq \Phi$}{
    $s$ element of $S$\;
    \If{$s$ is visited}{remove $s$ from $S$\;continue}
    \For{$d \in D $}
        {N $\longleftarrow$ $\Omega(s)$ \Comment*[r]{Neighbour finding function} 
        singular vecs $\longleftarrow$SVD(N)\;
        %Dominant vecs $\longleftarrow$ find dominant vecs (singular vecs) \;
        \If{SVM(Dominant vecs) is true}{
            $point$ $\longleftarrow$ $s$ + singular vec \;
            Add $point$ to $S$
        }
        \lElse{End points$\longleftarrow$ End points $\cup$  s}
            remove $s$ from $S$
    }
    Mark points in N as visited.
}
}
\end{algorithm}

\SetKwComment{Comment}{/* }{ */}
\begin{algorithm}[hbt!]
%\LinesNumbered
\caption{Cut-point detection}\label{alg:cut_point_localization}
\KwData{Graph with vertices ${V}$ and edges ${E}$}
\KwResult{Cut-point pose ${C(x,y,z,\alpha,\beta,\gamma)}$}
    {$C \longleftarrow $ Cut Points \\
    $v \longleftarrow V + V_B ; V_B \in \{Bud\ Vertices\}$ \\
    $e \longleftarrow E + E_B ; E_B \in \{Bud\ Edges\}$\\
    $n \longleftarrow $ No. of buds to keep \Comment*[r]{Pruning rule}
    $g \longleftarrow \{graph(v,e)\}$ \Comment*[r]{Set of disjoint graphs}
    $G \longleftarrow $ \textbf{min span tree}(g)\\
    $R \longleftarrow r$ \Comment*[r]{Root nodes}
    \For{ $i \in G $ }
        {$L_i \longleftarrow $ Leaf nodes of $G_i$\\
        $all\_paths \longleftarrow $ $\textbf{find\_multiple\_paths}$($G_i$, $R_i$, $L_i$, $thresh$)\\
        $unique\_paths \longleftarrow $ $\textbf{sequence\_match}(all\_paths)$\\
        \For {$i \in unique\_paths$}
             {$bud\_index \longleftarrow $ $\textbf{depth\_first\_search}(unique\_paths_i)$ \\
             $P \longleftarrow$ $\{$$bud\_index[n]$, $bud\_index[n+1]$$\}$\\
             }
        $C \longleftarrow \{$ P.x, P.y, P.z, $\alpha , \beta, \gamma \}$
        }
    }
\end{algorithm}

%\begin{figure} [!h]
%    \centering
%    \includegraphics[width=0.5\linewidth]{images/algorithm2.png}
%    \hspace{.2in}
%    \caption{Algorithm 2}
%    \label{fig:algorithm2}
%\end{figure}
\subsubsection{Pruning rule}
Pruning rules define a systematic way to remove older canes to keep the vigor and vine balance in control. Cane pruning and spur pruning are two of the most practiced pruning strategies in the U.S. grape industry. One major difference between these two rules is the number of buds retained after pruning. Cane pruning usually retains longer cane segments with variable counts of buds per canes, while spur pruning leaves fixed but smaller number of buds per canes. Algorithmically, cane pruning requires the ability to track buds in longer sections of canes, which could be a very difficult task because of the high degree of entanglement between the canes. On the other hand, spur pruning has simpler requirements and pruning locations are close to the cordons. For a proof-of-concept robotic pruning of vines, we adopted a simplified spur pruning rule to only retain 4 buds per cane. In addition to bud retention, pruning rules also necessitate qualitative parameters such as cane diameter and health of canes and buds. In this proof-of-concept design, we considered all canes for pruning and list the inclusion of qualitative parameters as future enhancements.

\subsubsection{Cut-point detection} 
\label{sec::cut_point}
Once the points belonging to just canes are segmented from the rest of the vine structure, the next step in the pipeline was to identify pruning locations. One possible approach could be to further segment clusters of multiple canes into individual clusters and process each cane individually. However, an additional segmentation or clustering step has the potential to induce more uncertainties, as segmentation and clustering processes are not perfect in themselves. So our approach deviates from this logic and processes all segmented canes at once using a graph-based approach. The SVM model explained in the previous section not only labels cane regions in the 3D models, but intersection regions between canes with cordons as well. The algorithm described in this section essentially uses the cane-cordon intersection region to solve the graphs. 

\begin{figure} [h]
    \centering
    \includegraphics[height=2.5in]{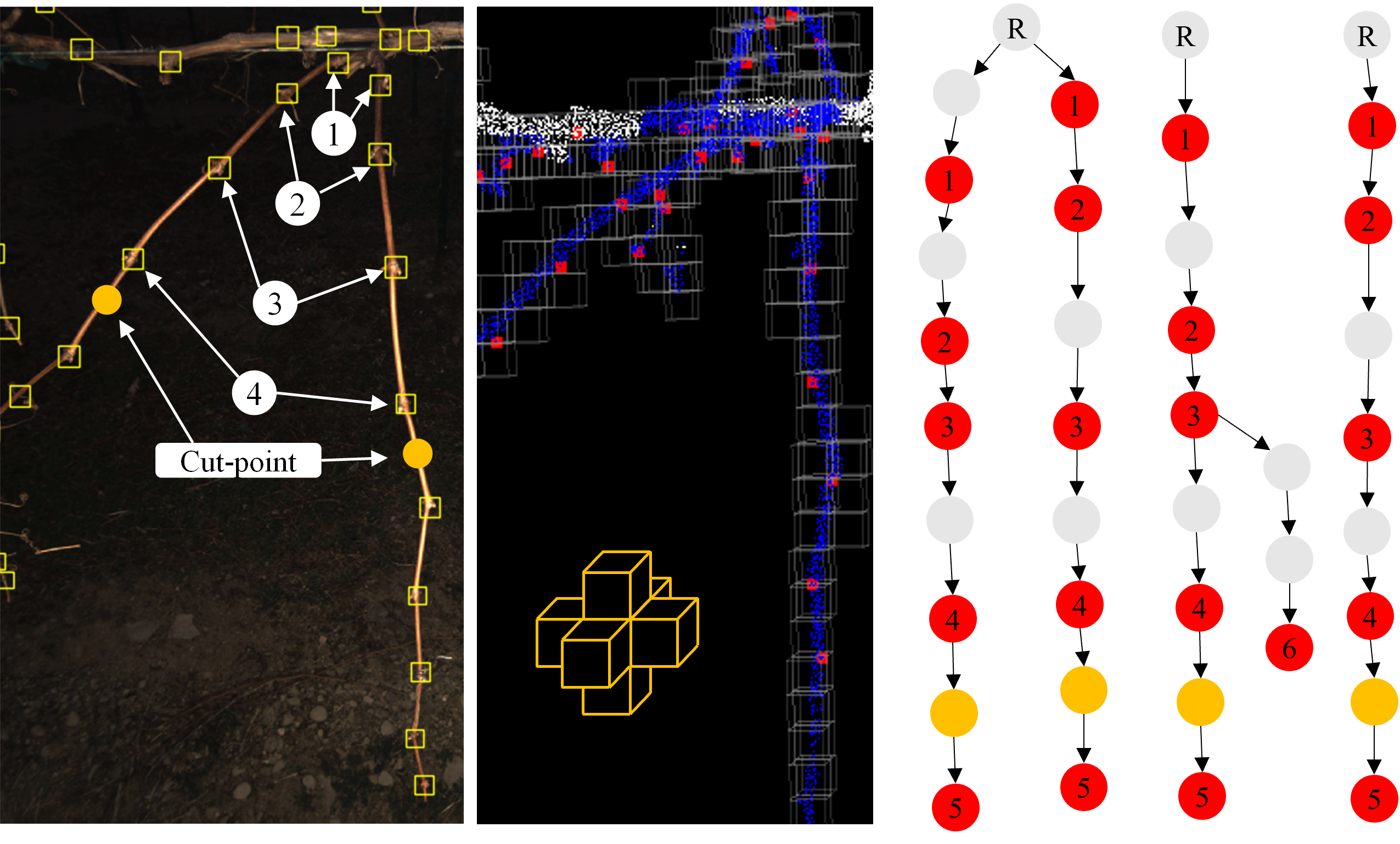}
    \hspace{.2in}

    \caption{Method employed for counting buds manually (left). Octree voxel representation of the point cloud model (center). Various examples of graph-based pruning location identification (right).}
    \label{fig:pruning_locations}
\end{figure}

To identify pruning locations, the foremost requirement in our pipeline was to convert the segmented cane-bud point  clouds into an undirected acyclic graph $G$. The first step in this process involved the use of Octree data structure to voxelize the cane clusters. The octree data structure recursively subdivides 3D point clouds into octants or voxels until a minimum voxel size is reached \cite{wurm2010octomap}. Here, an octree with a resolution of 5 cm was used to voxelize the extracted cane point cloud. Once the voxelization process was completed, the centroid of each voxel was extracted as a sub-sample of the canes. This was necessary to maintain the size of the graph and to keep computation time as low as possible. Subsequently, to generate the graph, a 3D kernel (Fig \ref{fig:pruning_locations}) traversed throughout the equally spaced octree centroids of the canes. With each step, the sixteen-neighbor kernel assigned vertices and edges to the voxelized cloud. The vertices and edges that contained buds were assigned to a special set of vertices $V_B$ and edges $E_B$. The post-processing of the graphs mainly involved the removal of loops using the minimum spanning tree (MST) algorithm. It is a greedy approach that removes cycles in weighted graphs while picking smallest weighed edges. To preserve the edges belonging to bud positions, the set of special edges $E_B$ were assigned smaller numerical weights, whereas the rest of the edges were initialized to larger weights. This simple step inherently preserved all bud vertices and edges, as the global cost of the graph was minimized. Figure \ref{fig:pruning_locations} demonstrates this logic.

Pruning rules require the correct ordering and numbering of buds per canes, which in our case requires to properly assign each bud to its respective location in the canes. This process essentially involved converting the undirected graph to a tree-like data structure that assigned directions in the graph $G$. By assigning the cane-cordon intersection region as root nodes in the graph, a depth first search-based (DFS) algorithm first computed all possible paths to the leaf nodes. As canes have random and complex 3D structures with or without branching, the paths from the root node to the leaf nodes could have multiple overlapping routes (Fig. \ref{fig:pruning_locations}). To suppress this ambiguity, a similarity score was computed on all the paths generated by the DFS algorithm using the sequence matching algorithm from \cite{jeh2002simrank}. This similarity score, essentially quantified path overlaps between the root nodes and the end points. A threshold of 0.9 (90\% similarity) was used to discard redundant routes in cases if multiple routs were found between the root node and the leaf nodes on the same cane. On these unique paths, newly discovered buds with respect to the root node were sequentially ordered, and the pruning points were identified using the pruning rule described in the previous Section. In our approach, the reduced canopy complexity with pre-pruning greatly facilitated this heuristic-based bud association algorithm. However, in complex vines with multiple crisscrossing canes, a more robust approach might be necessary.  

\begin{figure} [h]
    \centering
    \includegraphics[height={0.4\linewidth}]{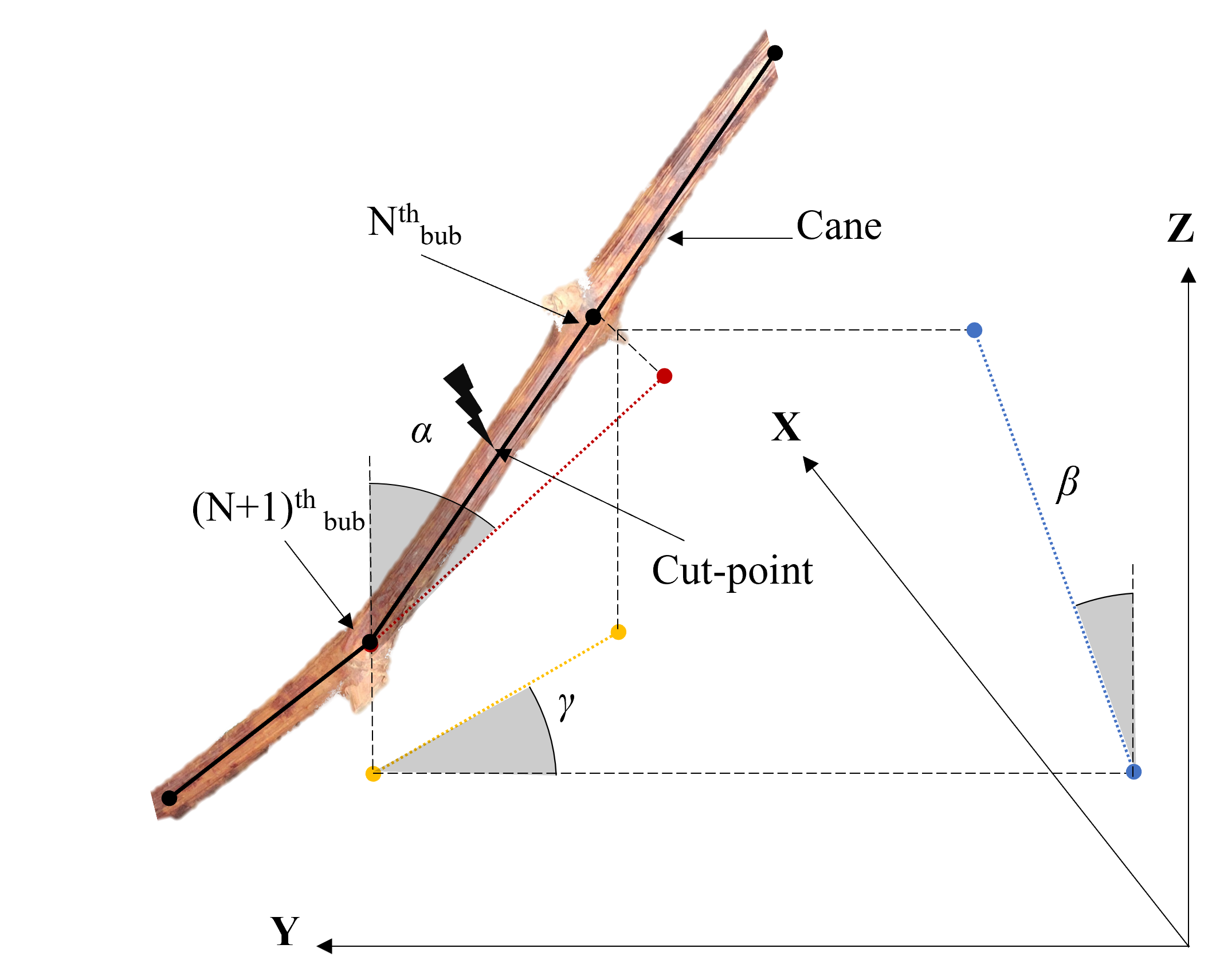}
    \hspace{.2in}
    \includegraphics[height=2.5in]{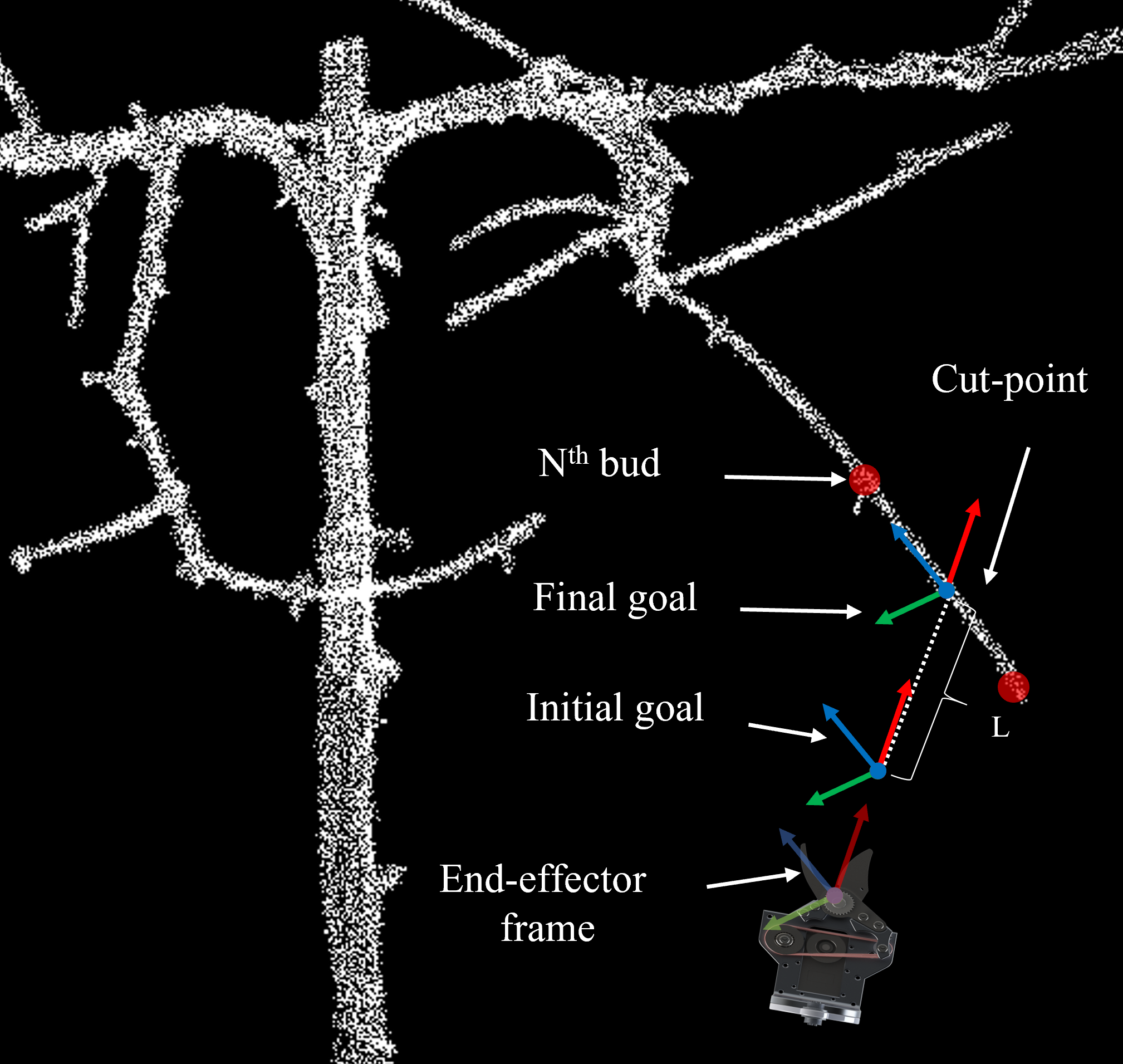}
    \caption{Cut-point detection (left). Three-dimensional vector projection of a cane section in the YZ (red line – roll angle ($\alpha$)), ZX (blue line – pitch angle ($\beta$)) and XY axis (yellow line – yaw angle ($\gamma$)). Pose of the angle made by the projected line made with respective axes. Initial and final approach to the pruning location (right).}
    \label{fig:cut_point_id}
\end{figure}

Once the pruning locations were identified, the next step in the pipeline was to compute its pose (position and orientation). A full pose was required as the cutting tool needs to approach the bud with a certain orientation to successfully make the cut. To calculate the cut-point orientation, we projected cane segments as 3D vectors on all three perpendicular planes. Here a 3D vector is defined by the 3D coordinates of the $N^{th}$ and $(N+1)^{th}$ bud section of the each cane that are projected to the XY, YZ, and ZX plane. The angles made by the projected line to the respective planes then provide the roll, pitch and yaw angles with respect to the reference frame. The mid-point of the vector was taken as the pruning location. This process is depicted in Fig. \ref{fig:cut_point_id} (left). 

\subsection{Manipulation}
\subsubsection{Motion planning}
\label{sec::planning}
The kinematic redundancy offered by the 7-DoF manipulator only reaches its full potential when the motion planning algorithm can incorporate all degrees of freedom in its planning context. The mechanical design of the robot (Fig. \ref{fig:robot_workspace} left) uses a 6 DoF off the shelf robot arm with a custom-built prismatic base. With few modifications, the ROS-MoveIt based planner \cite{chitta2012moveit} can be redesigned to plan the joint trajectories for an integrated 7 DoF arm, but at hardware level custom software drivers split the trajectories in real time and control all joints synchronously. To plan motion between all cut-points, we choose RRT-Connect \cite{kuffner2000rrt} as an option in the OMPL integrated with ROS that utilizes the entire 7 DoF. The RRT-Connect motion planner provides collision free motion planning  features which greatly fit the requirements for this proof-of-concept prototype (e.g., real time operation, obstacle avoidance or constrained movement). Additionally, in a comparative study of motion planners for robotic pruning by \cite{paulin2015comparison}, RRT and its variants were found to have overall better performance.

Once the poses of the pruning locations were computed, the path of the end-effector to the end goal positions were divided into to two discrete sets of trajectories.
The first set were computed using the RRT-connect solver and were used to make an initial approach to the cutting point, positioning the tool 15cm ahead of that cut-point (Fig. \ref{fig:cut_point_id} right). Once the tool was in this location, the end-effector was commanded to orientate perpendicularly to the branch that contains the pruning point. Subsequently, to accurately position the cane in-between the cutting blades, the motion planning scheme then switched to a cartesian path planner. This cartesian path planner (also from OMPL library) was constrained to maintain the orientation of the end-effector while inching slowly towards the final pose in a straight line. Then the blades closed and opened to mark the end of a successful cut operation. This process proved to be effective in much of the experimental cases; however, it is open-loop as the robot does not have any real-time feedback during the final approach to the cutting point. In Section \ref{sec::discuss}, we discuss some of the ways to close this loop to enhance the robustness of the system.

One way to define obstacles for motion planning could be to take the entire vine structure as obstacle and force planning algorithms to find solutions for all pruning locations. However, motion planning with collision detection and avoidance can be computationally expensive, especially in the unstructured and complex environment of dormant vines. The random arrangement of canes in the robot’s workspace as obstacles could result in the failure to converge to a solution or -as seen in practice- generate trajectories that result in erratic movements of the arm. To avoid such situations, collisions between the arm and the canes were allowed whereas trunk, trellis and cordon that are more structured and easier to identify were considered as obstacles. In addition to this, once a cutting action was executed, the arm always retracted backward to the initial pose (similar to the pose shown in Fig \ref{fig:robot_workspace} right) before planning the path to the next pruning location. This process generated not only a natural-looking motion, but most importantly provided more open space for the “connect” heuristics in the RRT-Connect planner to generate effective planning queries and solutions \cite{kuffner2000rrt}. 

Despite these systematic steps to supervise cautionary motion of the arm, contacts with canes would still be undesirable. To further minimize contact with canes, the task of sequencing the pruning locations for cutting operations was optimized using a travelling salesman problem (TSP). The nearest neighbor heuristic-based TSP exhaustively calculated all possible cutting route combinations and prioritized closer pruning locations to the ones farther from the end-effector while also minimizing the total travel distance. Although TSP is NP-harp problem, the small set of pruning locations per vine made the exhaustive optimization task possible in short duration. 

\subsection{Navigation}
\begin{figure} [!ht]
    \centering
    \includegraphics[height=2.7in]{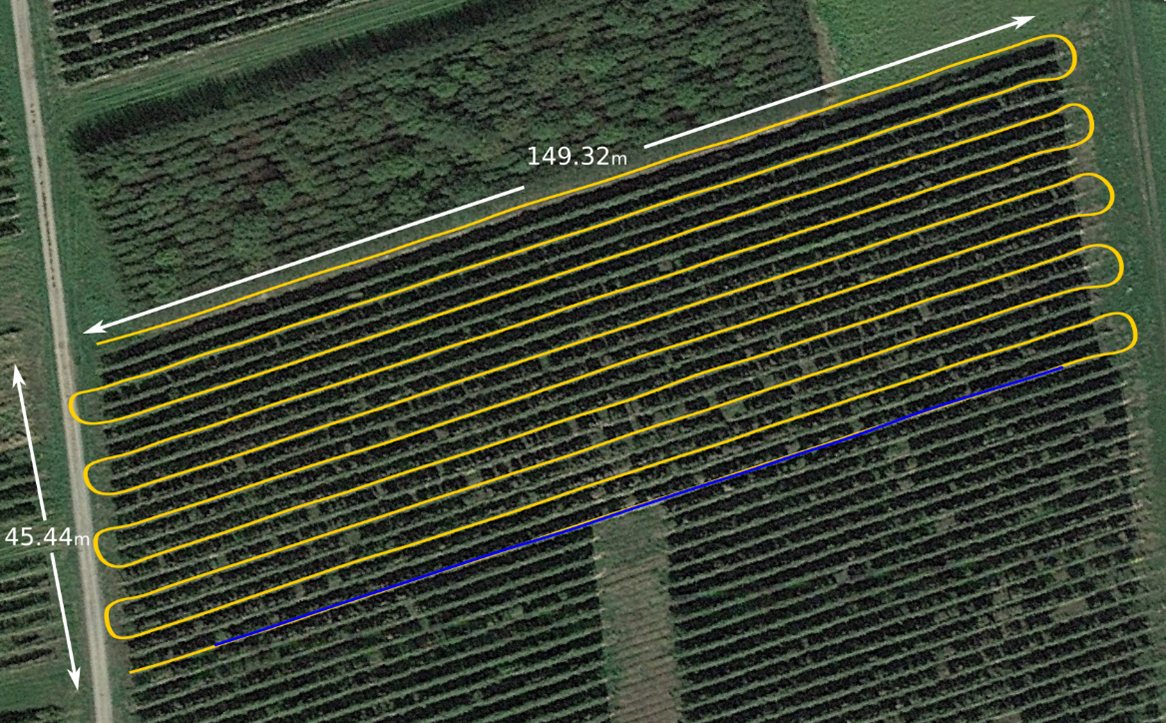}
    \hspace{.2in}

    \caption{Top view of the test site with RTK waypoint and pruning locations overlayed. The autonomous ground robot was capable to drive around the entire block without manual intervention (yellow path). The blue path correspond to the row selected for autonomous pruning.}
    \label{fig:vineyard_top_view}
\end{figure}

The  RTK-GPS receiver mounted on the vehicle, wheel encoders, and the robot’s onboard IMU (Inertial Measurement Unit) were the main localization sensors in the navigation system. An Extended Kalman Filter (EKF) fused the IMU, wheel odometry, and RTK position to localize the robot in the real world with the RTK-base as the reference frame. All RTK-GPS waypoints for autonomous navigation, including the locations of the vines  were manually collected. Also, all vines selected for pruning in this work were located in the same vineyard row (Fig. \ref{fig:vineyard_top_view}). However, the current architecture of the vine required us to address the same vine from both sides of the row. Thus, the main task for the navigation system was to drive the robot down the aisles, accurately turn and enter the aisle on the other side of the same row while stopping at each vine selected for pruning. To navigate in-between vineyard rows, we used a Model Predictive Controller (MPC) \cite{allgower2012nonlinear} to follow GPS waypoints.

The MPC controller first connected all waypoints using a spline to produce a smooth global path along with curvatures and speed profiles for the complete route. Subsequently, the RTK-GPS points corresponding to the pruning sites were read, and a local path from the current robot position to the next vine to be pruned was generated. Additionally, the planner also calculated local speed profiles (based on the global profile) along with the deceleration and acceleration required to smoothly and accurately stop in front of the tree trunk and start moving again. This approach allowed us to define parameters such as acceleration and deceleration ramps or cruise velocity, ensuring to provide a complete plan that minimized jerky motion that could potentially damage the robot’s components or the crops. For a system with state $x$ and control input $u$ at time $t$, the general discrete form of the MPC controller used is shown in Eqn. \ref{eqn::mpc}.
\begin{eqnarray} 
\label{eqn::mpc}
\displaystyle \min_{x(\cdot),u(\cdot)} & & \mathbf{x}^{T}(t_0+N)Q_f\mathbf{x}(t_0+N) + \sum_{t=t_0}^{t_0 + N} \mathbf{x}^{T}(t)Q \mathbf{x}(t) + \mathbf{u}^{T}(t)R \mathbf{u}(t) \nonumber\\
\textrm{subject to:} & & \mathbf{x}(t_0) = \mathbf{x}_0  \nonumber\\
\forall t \in \left[t_0, t_0 + N\right] : & & \mathbf{x}(t+1) = f( t, \mathbf{x}(t), \mathbf{u}(t) )\\
\forall t \in \left[t_0, t_0 + N\right] : & & 0 \geq s( t, \mathbf{x}(t), \mathbf(t) ) \nonumber
\end{eqnarray} 

Here, $N$ is the time horizon, $f (\cdot)$ is the robot motion model, $s$ represents the path constraints and $Q, R, Q_f$ are the weighting symmetric and positive (semi-) definite matrices. The span of the navigation system for autonomous pruning only required to travel a distance of two rows (approx. 0.3 km or 0.19 miles) and it included lane following, stopping between vines, turning, and re-entering. To adequately evaluate the autonomous navigation system,  we tested the self-driving capability in the entire block (1 mile/ 1.6 km). In this larger navigation experiment, the robot skipped a row entered a new row with every turn without stopping, as depicted by the yellow path in Fig\ref{fig:vineyard_top_view}. Overall, during this experiment the robot navigated 10 rows and made 9 U-turns autonomously.

\subsection{Systems integration}
\label{sec::systems_int}
\subsubsection{Robot platform}   
\begin{figure} [!ht]
    \centering
    \includegraphics[height=2.5in]{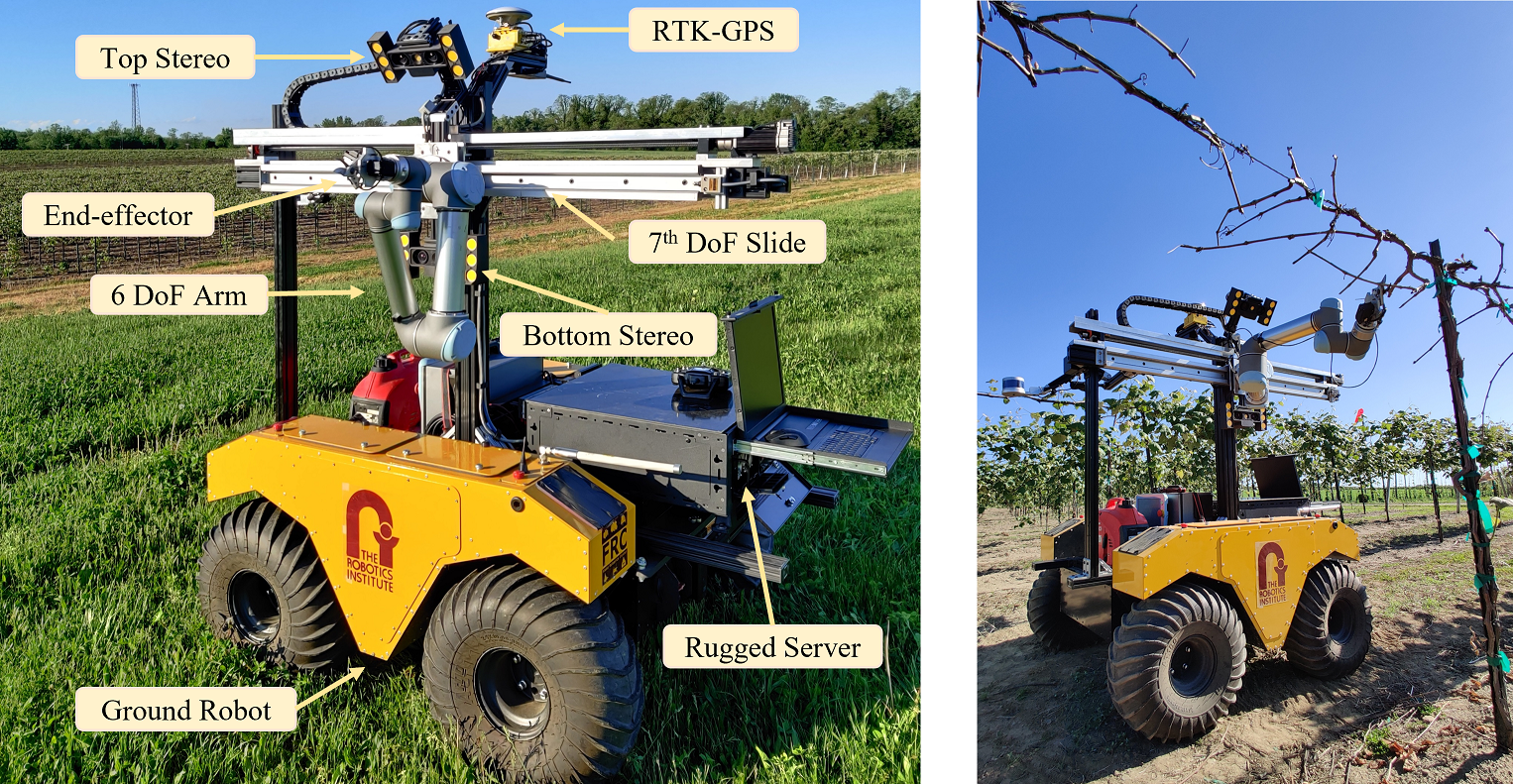}
    %\hspace{.2in}
    \caption{Integrated robotic system with 7 DoF robot arm, ground robot, cutting end-effector, dual stereo cameras, and on-board computers (left). An instance of autonomous pruning (right)}
    \label{fig:robot_system}
\end{figure}

The rugged ground robot (Warthog, Clearpath Robotics Inc.) fitted with a custom aluminum extrusion frame provided a base platform for the gantry system and the field server. The standalone integrated system with all perception, manipulation, and navigation components, and hardware are shown in Fig. \ref{fig:robot_system}. All electrical components including the computers, RTK-GPS, and cameras were powered by the ground robot’s battery except for the arm that was powered with the portable 1000 Watt gas generator for the AC control box. The edge field server ran on an Intel Xeon E5-2687Wv4 processor with 32GB of RAM and an NVIDIA GeForce GTX1080 GPU for deep neural networks inferencing. All software was packaged for ROS Kinetic under Ubuntu 16.04 LTS 64-bit Linux environment. A local NTP server was used to sync the clock for all sensors and computers for accurate temporal operations in ROS.

\subsubsection{Full autonomy}
\begin{figure} [h]
    \centering
    \includegraphics[height=2.25in]{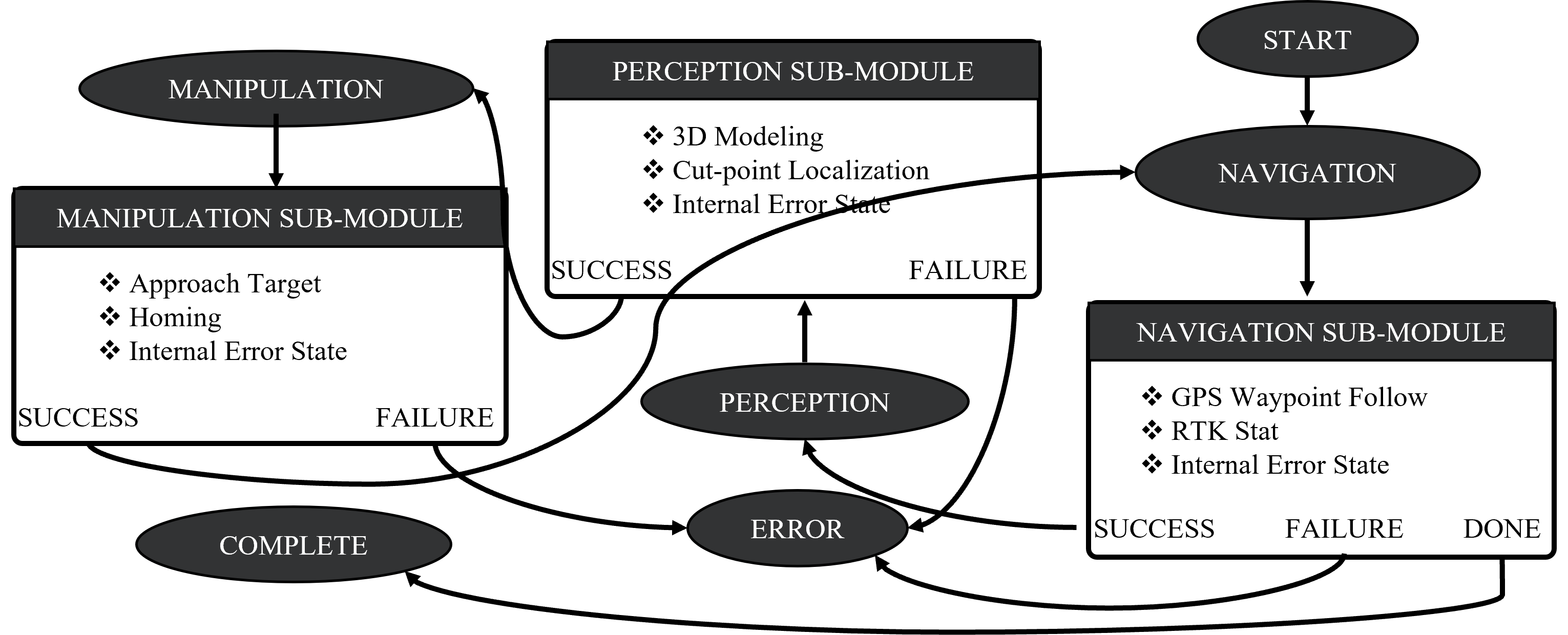}
    \hspace{.2in}

    \caption{Finite state machine. Depending on the state of the sub-modules, the state machine transitions between navigation, manipulation, perception, and error states for autonomous high-level control of the robot.}
    \label{fig:state_machine}
\end{figure}
One complete pruning cycle consists of several tasks that include navigating to the vine position, scanning and 3D modeling, identifying cut points and executing motion plans to physically removing canes from vines. For efficient high-level coordination and execution of multiple tasks, we used a finite state machine (FSM) as shown in Fig. \ref{fig:state_machine}. The states in the FSM were navigation, perception, manipulation, and error. Depending on the status of the sub-modules within each state, the SFM transitions between different states following a pre-defined sequence for autonomous high-level control of the robot until all vines were pruned. Additionally, for robustness, each of the sub-processes of the states were equipped with internal error sub-states to self-diagnose software level issues and pause all operations for manual intervention for hardware or unknown issues.

\section{Results}
\label{sec::results}
In order to evaluate all the systems that Bumblebee comprises, four datasets were employed. The first one was used to train and evaluate the bud detector. From this dataset, we selected randomly 5 samples (vines) to evaluate the reconstruction completeness and the region growing algorithm. The quality of the overall point cloud generated with the ICP approach was assessed using a single vine imaged in the field conditions. Finally, a total of 20 vines from a single row in a commercial vineyard were selected for pruning. These vines were pre-pruned with a mechanical pre-pruning machine to reduce the vigor and simplify the cluttered work environment. We also provide a brief analysis on how this non-selective process minimize the complexity of the vines, enabling our system to perform precise pruning.

The methodology employed for all these tests, as well as the results obtained are described in the following Sections. 
\subsection{Pre-pruning}
\label{sec::pre_prune}

\begin{wrapfigure}{r}{0.45\textwidth}
\centering
\includegraphics[width=0.4\textwidth]{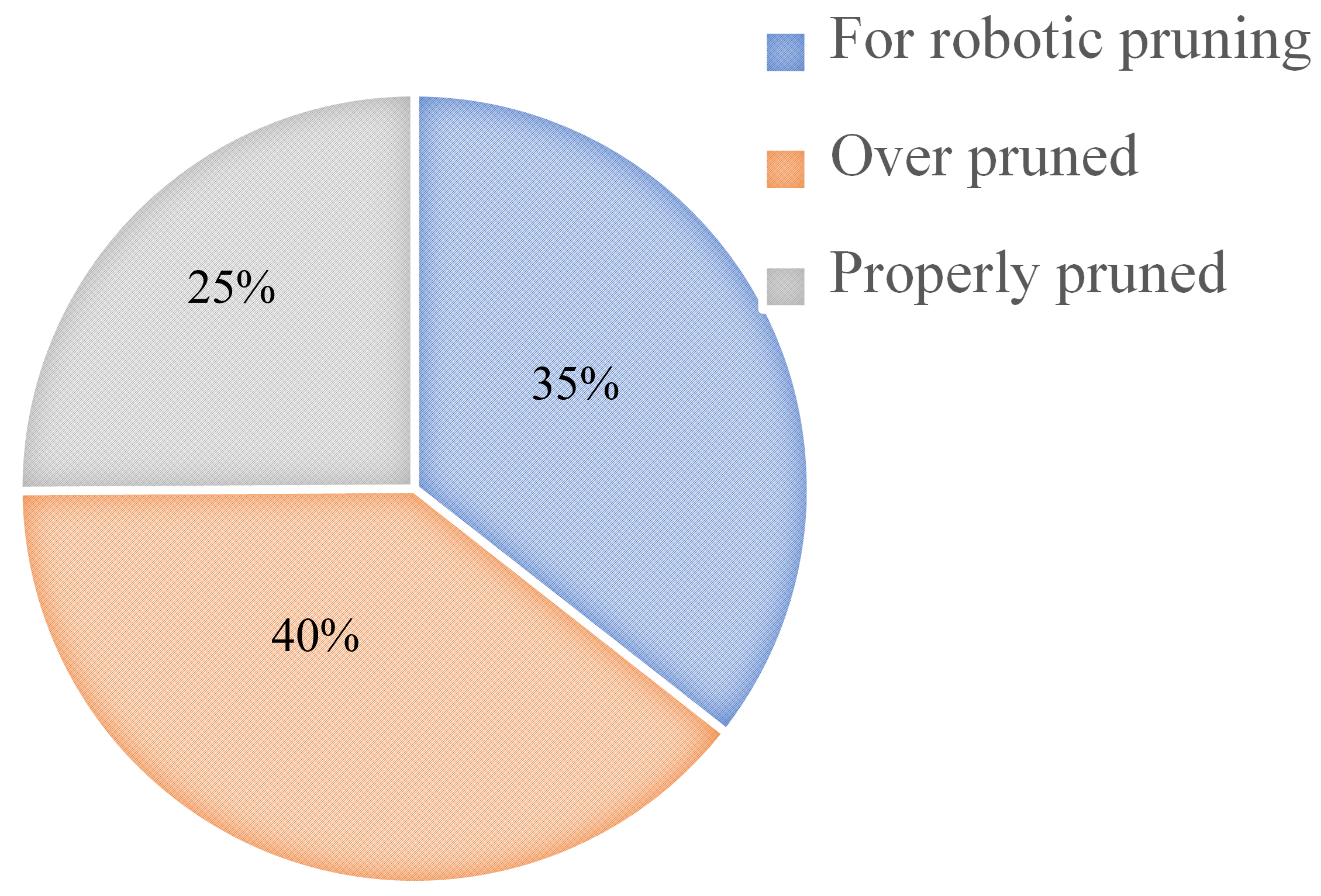}
\caption{Pre-pruning statistics.}
\label{fig:pi_chart}
\end{wrapfigure}

As detailed in Section \ref{sec::mods}, to simplify the canopy complexity, the vines were mechanically pre-pruned with an OXBO VMech 1210 Tool Arm and Sprawl pre-pruner (Fig. \ref{fig:pre-pruing} right). In the twenty field vines, we manually counted all canes as well as the number of buds per cane to evaluate the pre-pruning that set the stage for robotic operation. In total, 268 canes were present in these vines with an average of 13 canes per vine. After the pre-pruning operation, we observed that 25\% of the canes had exactly 4 buds, 35\% had more than 4, and 40\% were over pruned with less than 4 buds per cane. Here, ``4 buds'' is used as a reference as the simplified spur pruning rule adopted in this study only required to retain 4 buds per cane. Also, the distribution of buds ranged from 1 to 14 buds per cane as a result of non-selective manual pre-pruning operation. This large variation in the bud distribution is the variable we aim to minimize with our robotic pruning system, and the results are presented in the following subsections. As seen in Fig. \ref{fig:pi_chart} \& \ref{fig:Variability_plot}, the pre-pruning step not only greatly reduced the vigor of the vines and length of each cane, but also reduced the total number of canes to be pruned (35\%). Additional statistics include 1122 bud counts in total in all vines with a standard deviation of 2.08 bud counts about the mean of 4 buds per cane.

\subsection{Perception and Reachability}
Perceiving the environment is usually one of the earliest tasks for autonomous robots. In our case, detection of dormant buds in 2D images and their projection into the 3D coordinates along with the generation of the point cloud of the entire vine were some of the initial steps in the perception pipeline. As any inconsistency in bud detection or significant error in 3D reconstruction could highly affect subsequent processes such as estimating pruning locations and ultimately pruning, ensuring robust perception capabilities were crucial. 

\subsubsection {Camera system} 
\begin{figure} [!ht]
    \centering
    \includegraphics[width=0.95\linewidth]{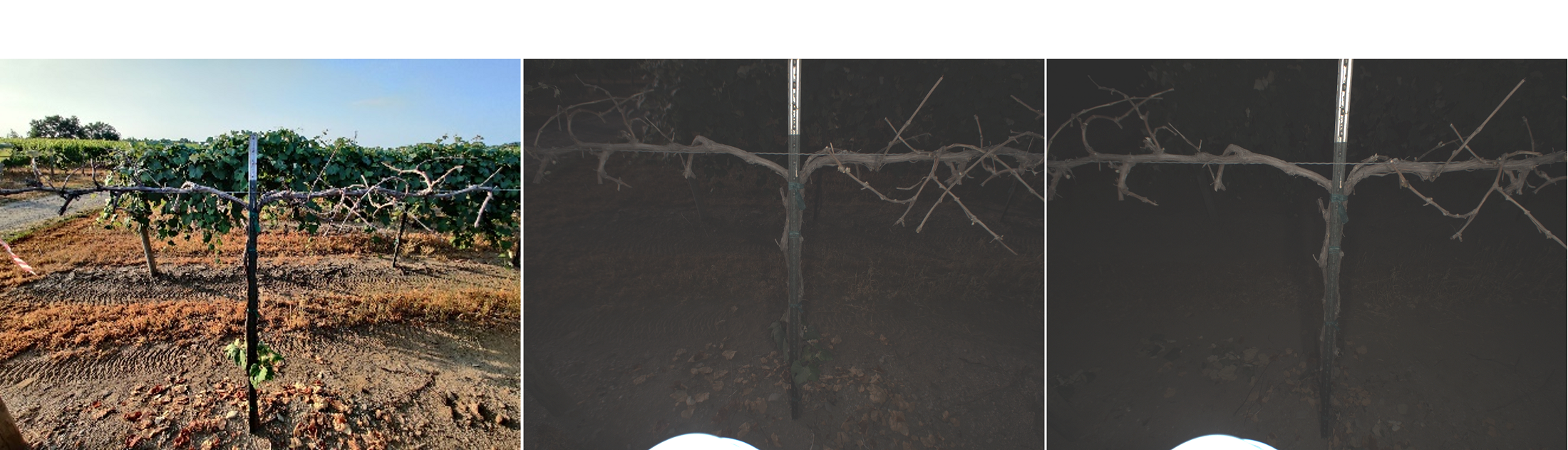}
    \hspace{.2in}
    \caption{Image of a test vine taken without flash (left). Image of the vine taken at the same time with flash (center). Night-time image of the same vine with flash (right).}
    \label{fig:image_quality}
\end{figure}

The active lighting camera described in Section \ref{sec::camera} was adequate to efficiently suppress affects from natural illumination. As a result, the camera system was able to produce images with consistent exposure (i.e., quality) in all lighting conditions present during the experiments. Figure \ref{fig:image_quality} shows images of the same vine taken at different time of the day with and without flash. The first image (Fig. \ref{fig:image_quality} left) was taken with the robot camera in a typical broad daylight whereas the remaining two were taken with the active light camera (Fig.\ref{fig:image_quality} (center) at same time as Fig. \ref{fig:image_quality} (left) and Fig. \ref{fig:image_quality} (right) at night-time). Qualitatively, Fig \ref{fig:image_quality} (center and right) look alike as the flash from the camera in conjunction with fast shutter speed overpowered day light effects. Quantitatively, the Structural Similarity Index Metric (SSIM) of images taken at various times of the day with the proposed camera system showed an average similarity of 90\%. The SSIM is a quality metric that embeds structural as well as contrast and luminance as quality parameters \cite{wang2004image}. 

\begin{figure} [h]
    \centering
    \includegraphics[height=1.9in]{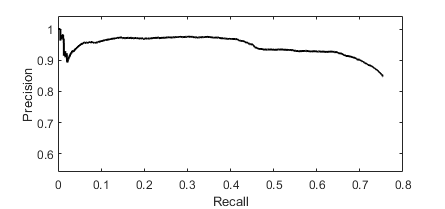}
    \hspace{.2in}
    \includegraphics[height=1.8in]{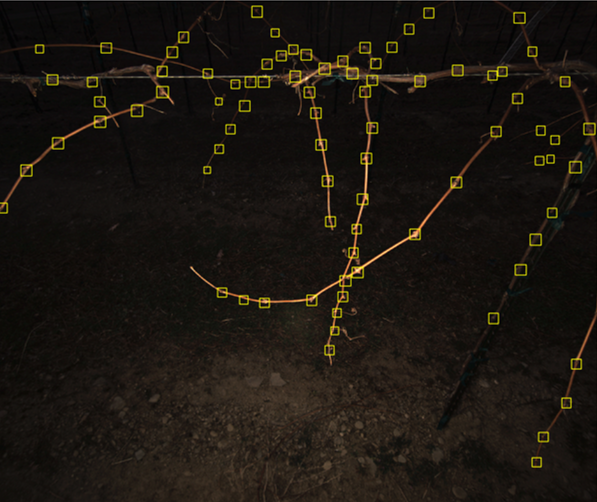}
    \caption{P-R curve for bud detection network (left). Each yellow bounding box localizes a bud in the image (right).}
    \label{fig:detection_metric}
\end{figure}

As mentioned in Section\ref{sec:bud_detection}, the bud detection network was trained on 85 images and evaluated in 35. Consistency in the mage quality  is considered as a major contributor for requiring such short amount of data to train our deep object detector. For instance, in Fig. \ref{fig:image_quality} (center and right), both images appear similar in exposure, color consistency, and background subtraction regardless of outdoor illumination. The P-R curve of the trained network is shown in Fig. \ref{fig:detection_metric}. Additionally, we obtained a mean precision average (mAP) of 0.93 in the test dataset. A thorough analysis of image quality from our camera system and reduction in training dataset size to fine-tune deep object detectors is explained in our previous publication (\cite{silwal2021robust}).

\subsubsection {Vine reconstruction}
To quantify the uncertainty in the measurement of depth information from cameras, depth measurements from stereo pairs were compared against highly precise and accurate laser measurements (±0.1 mm resolution). This process included comparing point stereo measurements from 30 different locations in a single vine at different depth against laser point measurements at the same location. The point measurements under comparison ranged from 0.2m up to 1m which also represented the reachable span of the robot arm. %This experiment showed that the custom stereo rig was able to get depth accuracy with an average uncertainty of ±3.7mm.
%\subsubsection {Reconstruction error}
As displayed in Fig. \ref{fig:robot_system}, the dual stereo rig that images each vine from 7 different positions,  producing 14 different views. The overall point cloud registration process was largely facilitated by the precise and accurate movement of the linear slider. As motion in all directions other than the slider was mechanically constrained, the initial estimates of the point clouds transformations prior to the ICP optimization were very accurate. With such initial estimates, the point to plane ICP optimally computed transformation between successive point clouds as well as the final point cloud registration described by Eqn. \ref{eqn:icp_linear} and \ref{eqn:icp_top_bottom}. Here, the ICP registration error is defined as the absolute difference between the ICP optimized translation against the actual distance travelled by the camera while imaging at different positions (measured with the encoder of the motor drive of the linear slide). The mean absolute error between these measurements averaged to ±2mm. %Similar to the stereo accuracy tests described before (for accuracy of depth measurement), 30 random vine locations from the full reconstructed 3D model were measured against the laser measurements. The measurements for the model sample points were taken by averaging the x, y and z location of neighboring point clouds. 
With uncertainties from individual stereo measurements and multiple ICP registration steps, the final accumulated registration error was estimated to be within ±6.8 mm. As explained in Section \ref{sec::ee}, the average diameter of the canes was 8 mm and the widest blade opening was 38 mm (Fig. \ref{fig:mechanical_design_end_effector}). Therefore, the accuracy achievable from the 3d reconstruction pipeline was well within the tolerance of the end-effector. All accuracy analysis were done in a mock-up vine in a laboratory setup. This was necessary mainly to rule out effects from wind that could alter both ground truth and the stereo measurements. All  laboratory tests used the same camera system as in the field prototype and a real vine collected from the test site. 

\subsubsection {Reconstruction completeness}
In addition to measuring reconstruction accuracy, the completeness of the 3D model also plays a crucial role in the overall success of autonomous pruning. For instance, largely fragmented cane structures and missing buds in the model could highly impact the cut point detection algorithm (see Section \ref{sec::cut_point}), which ultimately affects the pruning efficiency. In the literature, the quality of point clouds are usually assessed using objective and subjective metrics \cite{karantanellis2020evaluating}. Objective metrics usually compare point clouds to a reference or well-defined objects in the scene \cite{karantanellis2020evaluating,moon2019comparison,zhang2018patch}.  Whereas, subjective evaluation are based on visual inspection and usually  involve completeness, density etc. as factors in point cloud assessment \cite{karantanellis2020evaluating}. However,  because of lack of consistent structures in vines, reference or ground truth point clouds are difficult to generate and are not available for comprehensive comparison. To assess quality of point cloud from our dual stereo camera system, we present the following objective metrics.

\begin{itemize}[leftmargin=*]
\itemsep0em
\item \textbf{Number of points}: \textit{Total number of points in the registered point cloud.} 
\item \textbf{Number of neighbors}: \textit{Average number of points within the search radius of a sphere with r = 0.05 m.}
\item \textbf{Surface roughness}: \textit{Average distance between each point in the point cloud to the best fitting plane using neighbors in the search radius of a sphere with r = 0.05 m.}
\item \textbf{Surface Density}: \textit{Average number of points per square meter.}
\item \textbf{Volume Density}: \textit{Average number of points per cubic meter.}
\end{itemize}

Furthermore, we compare the above objective metrics between three models: one reconstructed only using the bottom camera (BC), the other reconstructed only using the top camera (TC), and finally with the registered point cloud using both the top and bottom cameras (TBC) on 5 vines. The results are shown in Table \ref{tab:quality_metric}. As expected, the results show that TBC model has more points, neighbors and higher surface as well as higher volume density when compared to BC and TC models. However, the surface roughness of the TBC mostly remained average between the TC and BC point clouds.

\begin{table}[!ht]
\centering
\caption{Objective quality metrics for vine point clouds.}
\label{tab:quality_metric}
\scalebox{0.78}{
\begin{tabular}{cccccccccccccccc}
\rowcolor[HTML]{D0CECE} 
\cellcolor[HTML]{D0CECE} &
  \multicolumn{3}{c}{\cellcolor[HTML]{D0CECE}No. of Points} &
  \multicolumn{3}{c}{\cellcolor[HTML]{D0CECE}No. of Neighbours} &
  \multicolumn{3}{c}{\cellcolor[HTML]{D0CECE}Roughness (mm)} &
  \multicolumn{3}{c}{\cellcolor[HTML]{D0CECE}Density (\#/$m^2$)} &
  \multicolumn{3}{c}{\cellcolor[HTML]{D0CECE}Volume (\#/$m^3$)} \\
\rowcolor[HTML]{D0CECE} 
\multirow{-2}{*}{\cellcolor[HTML]{D0CECE}Vine} & TC     & BC     & TBC    & TC  & BC  & TBC & TC   & BC   & TBC  & TC     & BC     & TBC    & TC      & BC      & TBC       \\
\rowcolor[HTML]{F2F2F2} 
1                                       & 32,134 & 23,197 & 43,479 & 381 & 359 & 600 & 6.22 & 5.50 & 5.89 & 48,524 & 45,735 & 76,415 & 727,863 & 685,956 & 1,146,260 \\
2                                       & 32,522 & 32,996 & 51,553 & 461 & 441 & 743 & 6.85 & 6.64 & 6.36 & 58,757 & 56,147 & 94,568 & 881,357 & 842,255 & 1,418,513 \\
\rowcolor[HTML]{F2F2F2} 
3                                       & 35,969 & 37,344 & 57,036 & 404 & 382 & 654 & 6.51 & 6.34 & 6.42 & 51,488 & 48,609 & 83,290 & 772,314 & 729,102 & 1,249,345 \\
4                                       & 29,256 & 25,642 & 42,764 & 418 & 396 & 669 & 6.35 & 5.64 & 5.85 & 53,239 & 50,481 & 85,221 & 798,623 & 757,188 & 1,278,310 \\
\rowcolor[HTML]{F2F2F2} 
5                                       & 27,912 & 23,457 & 39,925 & 377 & 387 & 630 & 5.95 & 5.70 & 5.72 & 47,958 & 49,245 & 80,185 & 719,353 & 738,648 & 1,202,803
\end{tabular}
}
\end{table}

For subjective evaluation, here we define two quality metrics to quantify the subjective quality of the reconstructed vine structures. The first metric, connected components, attempts to quantify the completeness of the vine structure as a function of the octree graphs connectivity described in Section \ref{sec::cut_point}. This approach essentially exploits the connected components properties of graphical structures. For a 3D model without any significant gaps in the model, we would anticipate a single or very few connected components, whereas fragmented/incomplete reconstruction would result in large numbers of connected components.  The second metric, bud counts, involves the number of buds in the reconstructed model. Similar to the objective metrics in the above paragraph, we compare the results from the TC, BC, and the TBC model on 5 test vines in the subjective evaluation as well. The results show that the number of incompletely formed canes were significantly reduced in the TBC model when compared to the TC and the BC models (6 vs. 17 vs. 19 respectively). Similarly, the number of buds that were essential features for cane segmentation were also significantly present in the TBC model than in the TC and BC model (425 vs. 306 vs. 311 respectively). In total, the TBC model had Mean Absolute Percentage Error (MAPE) of 5.11\% in buds count  whereas the TP and BC has significant higher MAPE in bud counts of 25.23\% \& 23.37\% respectively when compared to manually counted ground truth. Likewise, in the $R^2$ correlation between the manual counts of the buds vs. the final registered bud counts was highest in the TBC to TC and BC. Table \ref{tab:bud_counts} summarizes all these details on 5 vines for the TBC, BC, and TC models. The high level of completeness in the reconstruction of vines is attributed to the additional (elevated and slanted) views of the canopy provided by the top camera. Evidently, while point clouds from the bottom multiple views provided the majority of the vine structure, the top camera data filled gaps in the most occluded regions. Some of the fragments/disconnected canes in the TBC model were mainly stray canes from the adjacent vines, as no two consecutive vines had a well-defined separation.  

\begin{table}[]
\centering
\caption{Statistics on ground truth bud counts and buds present on different point cloud models.}
\label{tab:bud_counts}
\begin{tabular}{cccccc}
\rowcolor[HTML]{D9D9D9} 
Vine       & Ground Truth  & TBC & TC  & BC  & Correlation                        \\
1          & 86  & 88  & 64  & 66  & \cellcolor[HTML]{D9D9D9}Ground Truth vs. TBC \\
2          & 54  & 59  & 47  & 50  & $R^2$ = 0.98                               \\
3          & 77  & 83  & 60  & 63  & \cellcolor[HTML]{D9D9D9}Ground Truth vs. TC  \\
4          & 81  & 80  & 56  & 56  & $R^2$ = 0.96                               \\
5          & 121 & 115 & 79  & 76  & \cellcolor[HTML]{D9D9D9}Ground Truth vs. BC  \\
Total      & 419 & 425 & 306 & 311 & $R^2$ = 0.88                               \\
(100-MAPE)\% &     & 94.89  & 74.76 & 76.62 &                                   
\end{tabular}
\end{table}

\subsubsection {Region growing}
The validation of the region growing-based point cloud segmentation algorithm has two parts; i) accuracy of the SVM classifier, and ii) the resulting overall accuracy of cane segmentation. For the first part, we measured the performance of the SVM as a binary classifier to classify SVD decomposed values into cane regions. To evaluate the first part, we manually selected 536 sample point from 5 different vines (273 canes vs, 263 non-cane regions). Out of the 536 random test samples, the SVM correctly classified cane/non cane regions with a F1 score of 0.97. 

\begin{figure} [!ht]
    \centering
    \includegraphics[width=\textwidth]{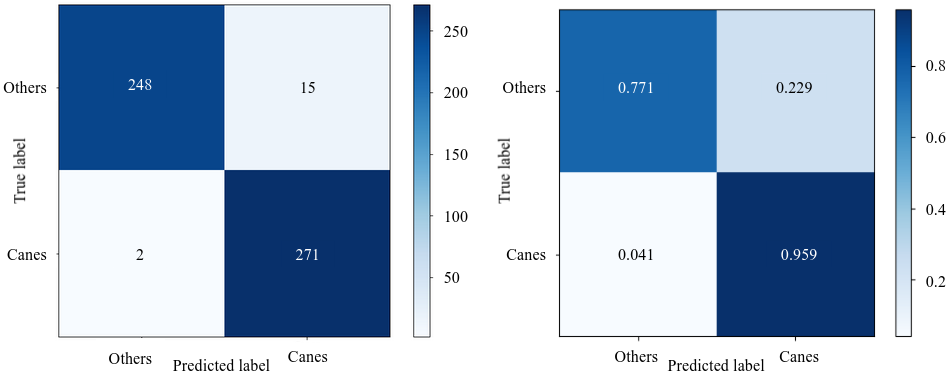}
    %\hspace{.2in}
    \caption{Confusion matrix for SVM-based cane/ non-cane classification on SVD values (left). Normalized confusion matrix for hand-labeled cane regions to region growing algorithm predictions (right). In both confusion matrices, the label "Others" refers to sample points not belonging to canes.}
    \label{fig:confusion}
\end{figure}

For the second part, we conducted  point-to-point comparison between the hand labeled cane point cloud to the region growing segmented point cloud. As hand labeling of complex vine structures are resource intensive, we are currently limiting the segmentation evaluation to 5 vines. The analysis shows that the cane segmentation pipeline achieved an overall F1 score of 0.91. The confusion matrix for the overall cane segmentation and SVM-based individual region classification are shown in Fig. \ref{fig:confusion}.
%Similarly, we have kept all the analysis in this section to 5 vines to remain consistent   

% For the second part, we compared the predicted 3D location of the cane-cordon intersection with the actual 3D location of cane-cordon intersections in the reconstructed model. As there is no benchmark to validate the cane-cordon intersection segmentation, we rely only on visual inspection of the algorithm labels to determine its performance at this time. Any SVM predictions that overlapped with the cane-cordon intersection region in the 3D model were classified as successful labels. Overall, the 

% For the final part, the accuracy of a cane segmentation was analyzed by comparing predicted cane point cloud labels with the hand labeled cane. The cane segmentation pipeline achieved an overall F1 score of 0.91. 

\subsubsection {Cut point localization}
To validate the localization accuracy of the cut points (in-between 4th and 5th bud), we kept track of all the canes (in the 20 vines) with bud counts exceeding 5. In all canes with a significant number of buds, the algorithm estimates of the cut positions were compared against the manual labels. Although, the midpoint between nth and (n+1)th bud was used as the 3D location of the cut points, any location within the two buds were taken as a valid solution since it was more important to correctly associate the bud sequencing. This algorithm on average achieved accuracy of 94\% across all manually selected pruning locations. 

\subsubsection {Workspace Quantification} 
\label{sec::reachability}
\begin{figure} [h!]
    \centering
    \includegraphics[width=\linewidth]{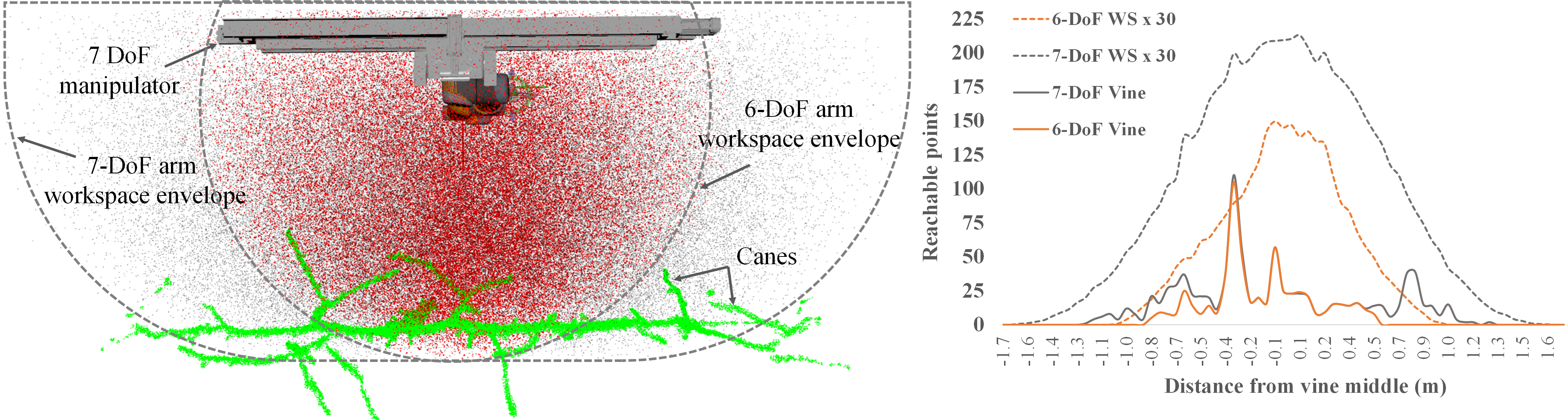}
    %\hspace{.2in}
    \caption{Top view of the work volume of the 6 and 7 DoF arms (left). The red and black dots are reachable positions for the 6\&7 DoF respectively. A graphical representation of the reachable locations in the vine for both 6 and 7 DoF arm (right).}
    \label{fig:reachability}
\end{figure}

A Monte Carlo experiment was carried out to estimate the volume of the robot's workspace. The idea was to sample points in the joint position space to estimate the overall reaching capability of the 7 DoF robot, for one vine. These tests were performed in a simulated environment, using a point cloud model of the field vines. In total, nearly 200,000  end-effector positions were  collected as samples of the reachable positions in the workspace. The volume enclosing all the positions reached by the end effector was estimated fitting a convex hull model implemented in MATLAB. This experiment was repeated for two cases: with the 6 DoF arm fixed to the center position of the linear slide and with the 7 DoF counterpart fully articulated allowing the prismatic base to move (Fig. \ref{fig:reachability} left).  As  expected, the results showed that the 3D work volume of the 7-DoF arm (3.5$m^3$) was more than 2 times higher compared to the lower 6-DoF design (1.6$m^3$). Similarly, because of the current architecture of the vines, this experiment also showed that if the mobile base is close enough to the canopy, an average of 68\% of the canes were within the reachable workspace of the manipulator,  while the remaining 32\% had to be addressed from the other side. A graphical representation of the number of points in the workspace and the number of reachable locations in the vine structure from the center of the vine is shown in Fig. \ref{fig:reachability}. In this figure (Fig. \ref{fig:reachability} left), the dashed lines represent the total reachable points in the work space, whereas the solid lines are the reachable points in the vine structure.

\subsection{Pruning}
\begin{figure} [h]
    \centering
    \includegraphics[height=2.75in]{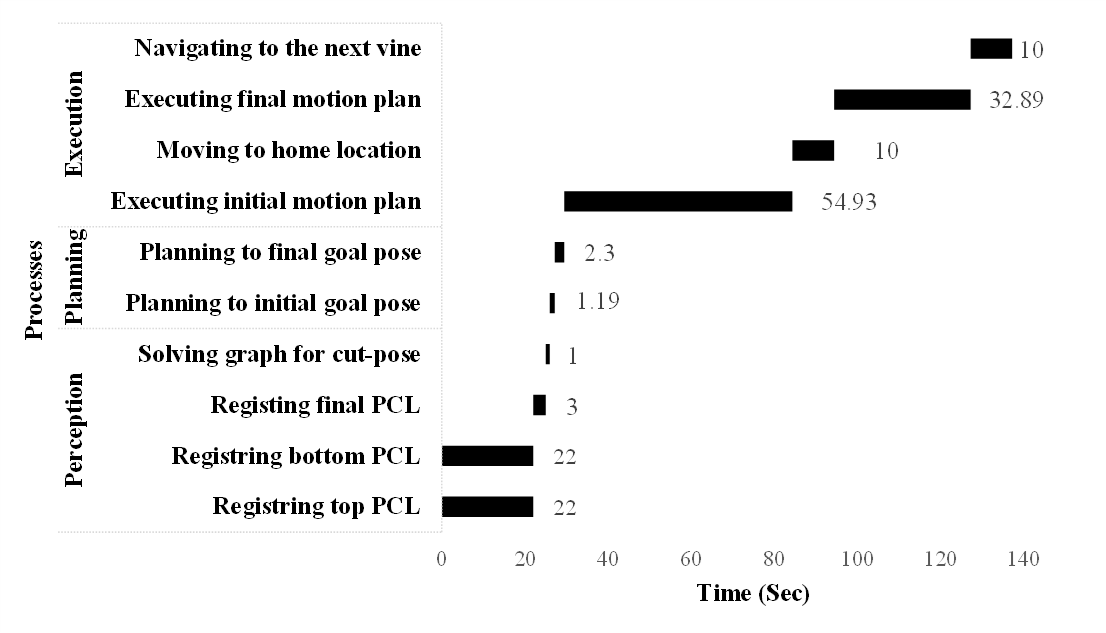}
    \hspace{.2in}

    \caption{Total computation timing breakdown.}
    \label{fig:comute_time}
\end{figure}

To measure the overall effectiveness of the presented robotic pruner, we introduce several metrics to evaluate its performance. First, Total Pruning Accuracy (TPA) quantifies the robot’s ability to prune successfully at the right pruning locations. Equation \ref{eqn:TPA} defines TPA as: 
\begin{equation}
\label{eqn:TPA}
    TPA\ =\ \frac{Total\ valid\ cuts}{Total\ pruning\ locations\ }\ 
\end{equation}
Similarly, Total Pruning Cycle (TPC) is the average time required to prune each vine, as described in Eqn. \ref{eqn:TPC}. This metric linearly combines computation cost of all sub-processes in the perception, planing, manipulation, and navigation systems. The computation timing breakdown of all major sub-operations in TPC is shown in Fig. \ref{fig:comute_time}. 
\begin{equation}
\label{eqn:TPC}
    TPC  = T_{perception} + T_{planning} + T_{execution} 
\end{equation}

Fig. \ref{fig:comute_time} shows the TPC from a single side pruning and the significant sub-processes under perception, planning, and execution stages. In total, it took 137 sec to prune a vine from one side. The current vine training system allowed canes to be randomly distributed on both sides of the canopy and nearly 32\% of the canes were on the opposite side and outside the reachable workspace of the robot. For this reason, the robot had to repeat all operation from both sides of the canopy (i.e., from point cloud model generation to motion planning and execution), which increased the TPC to 213 sec/ vine.

\begin{figure} [h]
    \centering
    \includegraphics[width=0.85\linewidth]{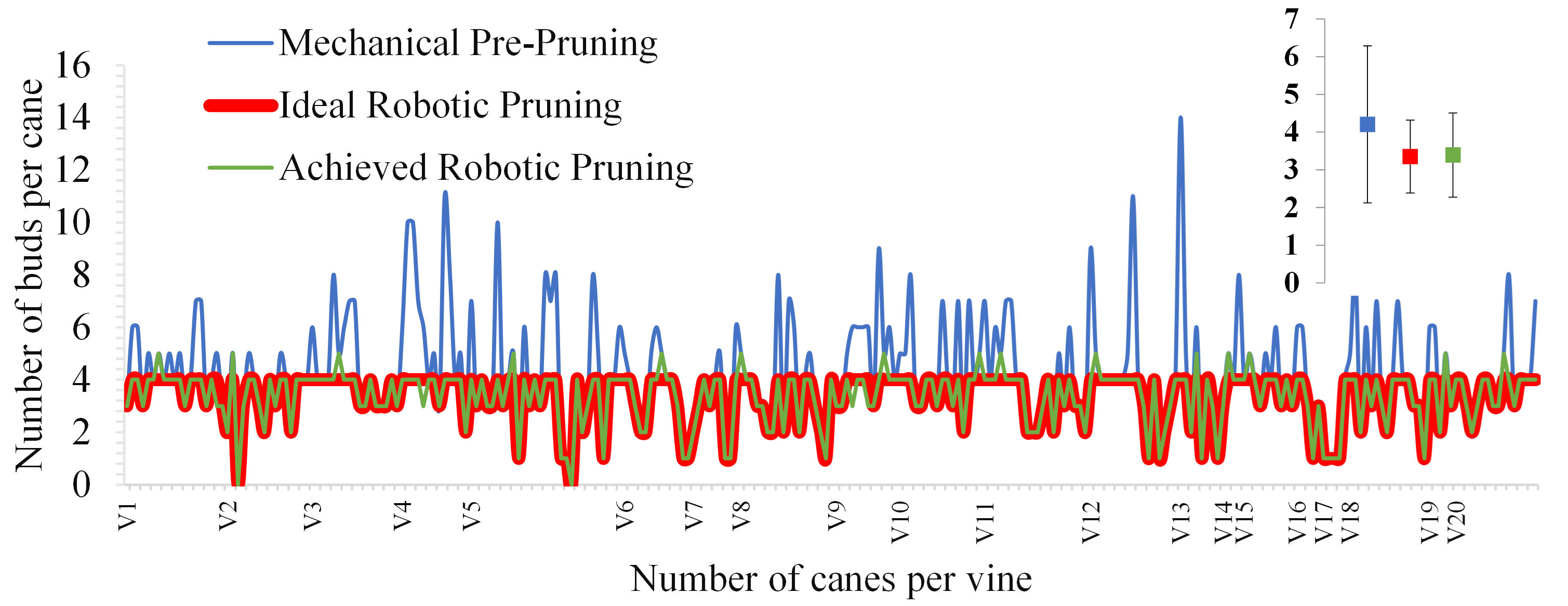}
    \hspace{.2in}
    \caption{A variability plot that shows the distribution of buds per cane for all field vines after pre-pruning, under the assumption of ideal robotic pruning, and achieved results with the robotic pruner. The error plot on the right side shows the standard deviation of bud distribution about the mean buds count.}
    \label{fig:Variability_plot}
\end{figure}

The variability caused by non-selective pre-pruning is shown in Fig. \ref{fig:Variability_plot}. After this operation, the standard deviation of the bud distribution per cane was found to be ±2.08. Based on the statistics from Section \ref{sec::pre_prune} (blue data lines), only 95 out of 268 canes (35\%) need to be pruned. Under the assumption of ideal perception and manipulation capabilities where all pruning locations were detected and pruned, the best achievable deviation would be of 0.97 standard deviation (Fig . \ref{fig:Variability_plot} red data line). In reality, because of some discrepancies in the pruning point detection and motion planning/ execution pipelines, not all canes were consistently pruned. Out of 95 prunable canes, only 83 were successfully pruned, yielding a TPA of 87\% (Fig. \ref{fig:Variability_plot} green data line). However, even with 87\% TPA, the standard deviation decreased to ±1.03 which is a significant reduction in variance given that the pre-pruning step over-pruned 45\% of the prunable canes. In section \ref{sec::discuss} we further discuss on the source and potential improvements of the current system.

The factor that ultimately determines the success criteria for a pruning robot is its ability to remove canes. In other words, all the steps in the perception and motion planning leading up to the final execution of the cutting action becomes significant only if the target cut-point gets successfully cut. Commonly used metrics such as the TPA described above only quantify the ratio of success or failure in completing the pruning tasks. To incorporate the effects of intermediate steps and to better describe the overall performance of the pruning robot, we introduce a new metric called Total Pruning Efficiency (TPE). The TPE is a multiplicative combination of several efficiency terms that at high level include perception (3D registration, cut-point detection), motion planning, and execution efficiencies as shown in Eqn. \ref{eqn:TPE}. In this metric, all efficiencies are converted accuracies or success rates normalized to a number between 0 and 1. For instance, the bud detection efficiency is essentially the accuracy of detecting buds where $\eta_{bud\ detection} = 0.95$ represents 95 \% detection accuracy compared to ground truth values (see Table \ref{tab:bud_counts}). Similarly,  $\eta_{planning} = 0.95$ represents 95\% success rate in the motion planner's convergence to a solution. As shown in Table \ref{tab:tech_combination}, the TPE is especially valuable in narrowing down system bottlenecks and as well as to justify different design choices.
\begin{align}
\label{eqn:TPE}
\textstyle TPE &= \textstyle \eta_{registration}\ * \textstyle \eta_{localization}\ * \textstyle \eta_{execution}\\
\textstyle \eta_{registration} &= \textstyle \eta_{bud\ detection}\ * \textstyle \eta_{3D\ reconstruction}\nonumber\\
\textstyle \eta_{localization} &= \textstyle \eta_{\ cane\ segmentation}\ * \textstyle \eta_{cut-point\ identification}\nonumber\\
\textstyle \eta_{execution} &= \textstyle \eta_{planning}\ * \textstyle \eta_{execute}\nonumber
\end{align}

Table \ref{tab:tech_combination} summarizes TPE for various possible combinations of hardware, software, and pruning strategies. The first row with the cell labeled “all inclusive”, incorporates all system components described in this paper. Here, the TPE accounted to 0.64 even with higher accuracies in the registration and localization pipeline but with relative low manipulator execution efficiency. In single side pruning, we only considered pruning a vine from one side. Although the $\eta_{registration}$ and $\eta_{localization}$ efficiencies remained similar, pruning only from one side mainly affected $\eta_{execute}$ efficiency as nearly 32\% of pruning location were out of reach. This decreased the TPE significant to 0.3. Considering the full point cloud model of vines as obstacle mainly affected the motion planning and execution efficiencies. With more occupied space in the robot’s workspace, the sampling-based planner (RRT-connect) took significant amount of time as well as attempts to converge to a solution. Significant delays were also observed in the motion execution, and most joint configurations looked unnatural and complex. The TPE in this case was 0.47 with TPC of 177. With just one stereo pair, we observed the most drastic effect to TPE. As it affected the perception part at the beginning of the pruning cycle, the error propagated to localization and motion execution stages. Here, the TPE was only 0.17. Finally, nearest neighbor-based TSP optimization exhaustively minimized the total point-to-point distance travelled while visiting all pruning locations. Without TSP, the longest possible cut-routes increased the TPC by nearly 20\% while the rest of the efficiencies remained similar. The single side, full vine model as obstacle, model from single stereo, and the pruning point sequencing without TSP were analyzed in a simulator, virtually pruned from single side, and based on the data of the same vines collected in this study.

\begin{table}[!ht]
\centering
\caption{\label{tab:tech_combination} Combination of different design choices that affect TPE.}
\scalebox{0.9}{
\begin{tabular}{ccccccccc}
\rowcolor[HTML]{C0C0C0} 
Combination &
  \multicolumn{2}{c}{\cellcolor[HTML]{C0C0C0} $\eta_{registration}$} &
  \multicolumn{2}{c}{\cellcolor[HTML]{C0C0C0} $\eta_{localization}$} &
  \multicolumn{2}{c}{\cellcolor[HTML]{C0C0C0} $\eta_{execution}$} &
  TPE &
  TPC \\
\rowcolor[HTML]{EFEFEF} 
\cellcolor[HTML]{EFEFEF} &
  $\eta_{bud \ detection}$ &
  0.95 &
  $\eta_{cane \ segmentation}$ &
  0.87 &
  $\eta_{planning}$ &
  0.95 &
  \cellcolor[HTML]{EFEFEF} &
  \cellcolor[HTML]{EFEFEF} \\
\rowcolor[HTML]{EFEFEF} 
\multirow{-2}{*}{\cellcolor[HTML]{EFEFEF}All inclusive} &
  $\eta_{3D \ reconstruction}$ &
  0.96 &
  $\eta_{cut-point  \  identification}$ &
  0.97 &
  $\eta_{execute}$ &
  0.84 &
  \multirow{-2}{*}{\cellcolor[HTML]{EFEFEF}0.61} &
  \multirow{-2}{*}{\cellcolor[HTML]{EFEFEF}245} \\
 &
  $\eta_{bud \ detection}$ &
  0.95 &
  $\eta_{cane \ segmentation}$ &
  0.87 &
  $\eta_{planning}$ &
  0.67 &
   &
   \\
\multirow{-2}{*}{Single   side pruning} &
  $\eta_{3D \ reconstruction}$ &
  0.96 &
  $\eta_{cut-point  \  identification}$ &
  0.97 &
  $\eta_{execute}$ &
  0.59 &
  \multirow{-2}{*}{0.3} &
  \multirow{-2}{*}{137} \\
\rowcolor[HTML]{EFEFEF} 
\cellcolor[HTML]{EFEFEF} &
  $\eta_{bud \ detection}$ &
  0.95 &
  $\eta_{cane \ segmentation}$ &
  0.87 &
  $\eta_{planning}$ &
  0.84 &
  \cellcolor[HTML]{EFEFEF} &
  \cellcolor[HTML]{EFEFEF} \\
\rowcolor[HTML]{EFEFEF} 
\multirow{-2}{*}{\cellcolor[HTML]{EFEFEF}Full   vine obstacle} &
  $\eta_{3D \ reconstruction}$ &
  0.96 &
  $\eta_{cut-point  \  identification}$ &
  0.97 &
  $\eta_{execute}$ &
  0.73 &
  \multirow{-2}{*}{\cellcolor[HTML]{EFEFEF}0.47} &
  \multirow{-2}{*}{\cellcolor[HTML]{EFEFEF}177} \\
 &
  $\eta_{bud \ detection}$ &
  0.75 &
  $\eta_{cane \ segmentation}$ &
  0.64 &
  $\eta_{planning}$ &
  0.92 &
   &
   \\
\multirow{-2}{*}{Single   stereo} &
  $\eta_{3D \ reconstruction}$ &
  0.68 &
  $\eta_{cut-point  \  identification}$ &
  0.68 &
  $\eta_{execute}$ &
  0.84 &
  \multirow{-2}{*}{0.17} &
  \multirow{-2}{*}{105} \\
\rowcolor[HTML]{EFEFEF} 
\cellcolor[HTML]{EFEFEF} &
  $\eta_{bud \ detection}$ &
  0.95 &
  $\eta_{cane \ segmentation}$ &
  0.87 &
  $\eta_{planning}$ &
  0.92 &
  \cellcolor[HTML]{EFEFEF} &
  \cellcolor[HTML]{EFEFEF} \\
\rowcolor[HTML]{EFEFEF} 
\multirow{-2}{*}{\cellcolor[HTML]{EFEFEF}No-TSP} &
  $\eta_{3D \ reconstruction}$ &
  0.96 &
  $\eta_{cut-point  \  identification}$ &
  0.97 &
  $\eta_{execute}$ &
  0.84 &
  \multirow{-2}{*}{\cellcolor[HTML]{EFEFEF}0.59} &
  \multirow{-2}{*}{\cellcolor[HTML]{EFEFEF}164}
\end{tabular}
}
\end{table}

%\subsection{Computational cost breakdown}
%\begin{figure} [h]
%    \centering
%    \includegraphics[height=2.75in]{images/timing_plot.png}
%    \hspace{.2in}
%
%    \caption{Total computation timing breakdown.}
%    \label{fig:comute_time}
%\end{figure}

%Fig. \ref{fig:comute_time} shows the TPC from a single side pruning and the significant sub-processes under perception, planning, and execution stages. In total, it took 137 sec to prune a vine from one side. The current vine training system allowed canes to be randomly distributed on both sides of the canopy and nearly 32\% of the canes were on the opposite side and outside the reachable workspace of the robot. For this reason, the robot had to repeat all operation from both sides of the canopy (i.e., from point cloud model generation to motion planning and execution), which increased the TPC to 213 sec/ vine.

\subsection{Navigation}

\begin{figure} [!h]
    \centering
    \includegraphics[width=0.8\linewidth]{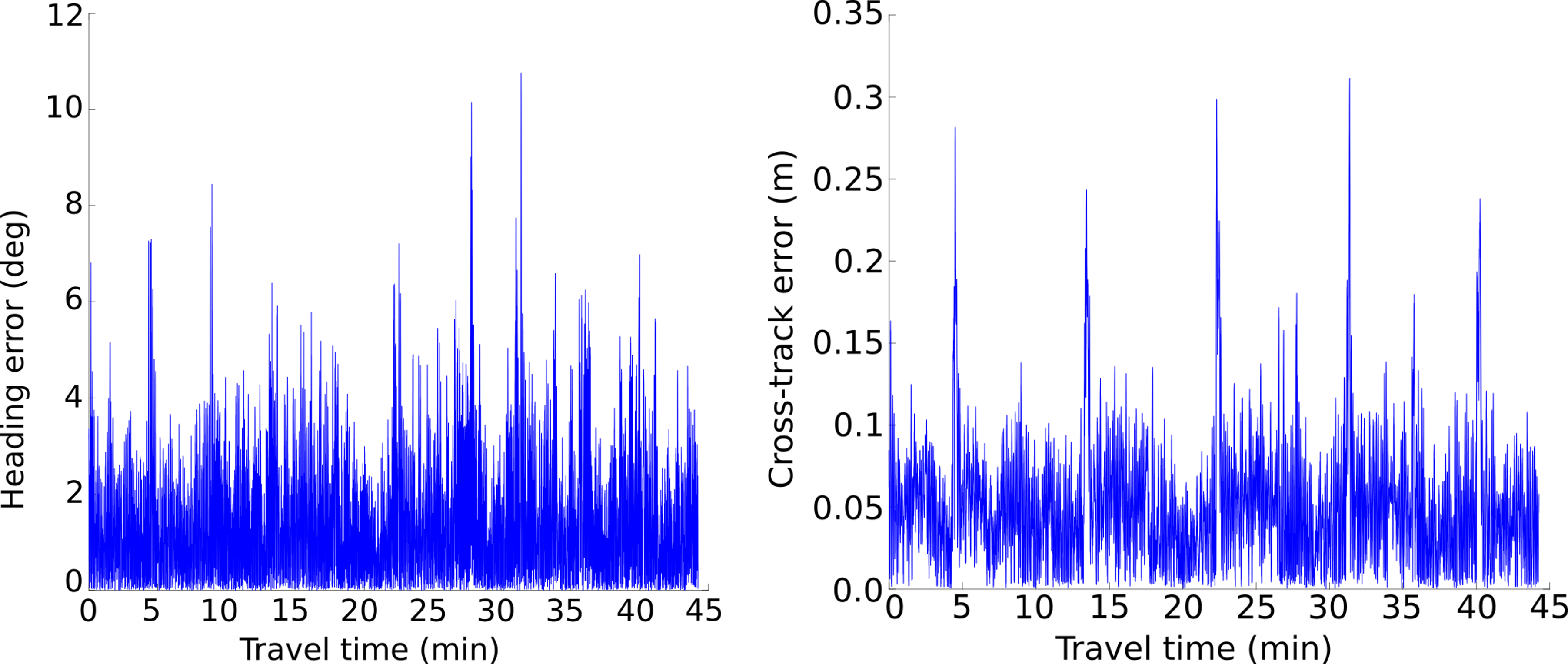}
    \hspace{.2in}

    \caption{Cross-track and heading errors of the path followed by the vehicle when compared with the desired trajectory generated by the planner.}
    \label{fig:nav_error}
\end{figure}

As mentioned in Section \ref{sec::systems_int}, the robot drove to each pruning location, remain stopped while pruning and start moving again to the next vine location. Given that the width of the vineyard row was only 1.8 m wide and the mobile robot with the linear slider and the arm was approximately 1.2 m, maintaining a consistent distance from the canopy and remaining parallel to the rows were critical requirements. The autonomous navigation system was first tested on the accuracy of stopping at each pruning location. We used 20 vines whose positions were marked prior to the trial using the RTK-GPS.  During the test, the robot stopped as expected in all locations and the average position of the robot while stopped was used to calculate its distance to the desired vine location. The longitudinal root mean squared error obtained in this case was 0.28 m, which was acceptable given the pruning task was accomplished successfully in all 20 vines. Laterally, we observed average cross-track errors of 0.07 m for in-row navigation and maximum deviation errors of 0.29 m, which mainly occurred while turning. The average heading angle error was 10.76 degrees and when stopped, the robot remained parallel to the canopy. This positioning facilitated the 3D reconstruction and motion planning algorithms for the pruning task, as all the pruning points were horizontally equidistant to the cameras for imaging and the arm for the actuation.

We also evaluated the capability of the autonomous navigation system in the larger section of the vineyard with multiple rows. To this aim, the robot was commanded to drive skipping one row along the yellow route depicted in Fig. \ref{fig:vineyard_top_view}. The total travel distance driven was approximately 1571.13 m at an average speed of 0.58 m/s and the robot drove autonomously for approximately 45 minutes with no intervention. In both trails, similar heading angle and cross tracking errors as in the pruning section were observed.

\section{Discussion}
\label{sec::discuss}
In general, robots that interact with their environment pose very challenging problems to solve. In particular for robotic pruning, biologically driven surrounding and indeterminant growth habits of vines add more challenges in perceiving  and interacting with the environment. It took us three years of effort with two hardware revisions and numerous software modifications to achieve the results reported in this paper. This section summarizes the key lessons learned, capabilities and limitations, future enhancements to our existing system and some remarks to guide further research. 

The first requirement in the perceptual capabilities of the pruning robot is accurate and complete 3D models of vines. With multiple views (fourteen different viewpoints), the scan-match based 3D reconstruction approach was able to generate precise models of the vines. In the generated 3D modes, canes which are thin structures with diameters as small as 4 mm were clearly visible with few fragments in its structure. The top slanted camera was a necessary addition, which greatly helped to minimize missing information in the occluded regions by adding point clouds from views that were not seen from the front facing camera. Thus, with adequate overlapping from multiple viewpoints, the complete 3D reconstruction of the vines was possible from just one side of the canopy. However, this approach not only required frequent stereo calibration but was also required manual tuning of several parameters to maintain a relatively consistent size of the registered point clouds for real-time processing. However, modern commercial vineyards typically have consistency in row width and vine spacing, and are equipped with mechanical means to simplify vine complexity at scale. These factors made it possible to tightly control field experiments such as maintaining constant distance between the robot and vines to achieve consistent results even with heuristically chosen parameters.

\begin{figure} [h]
    \centering
    \includegraphics[width=0.9\linewidth]{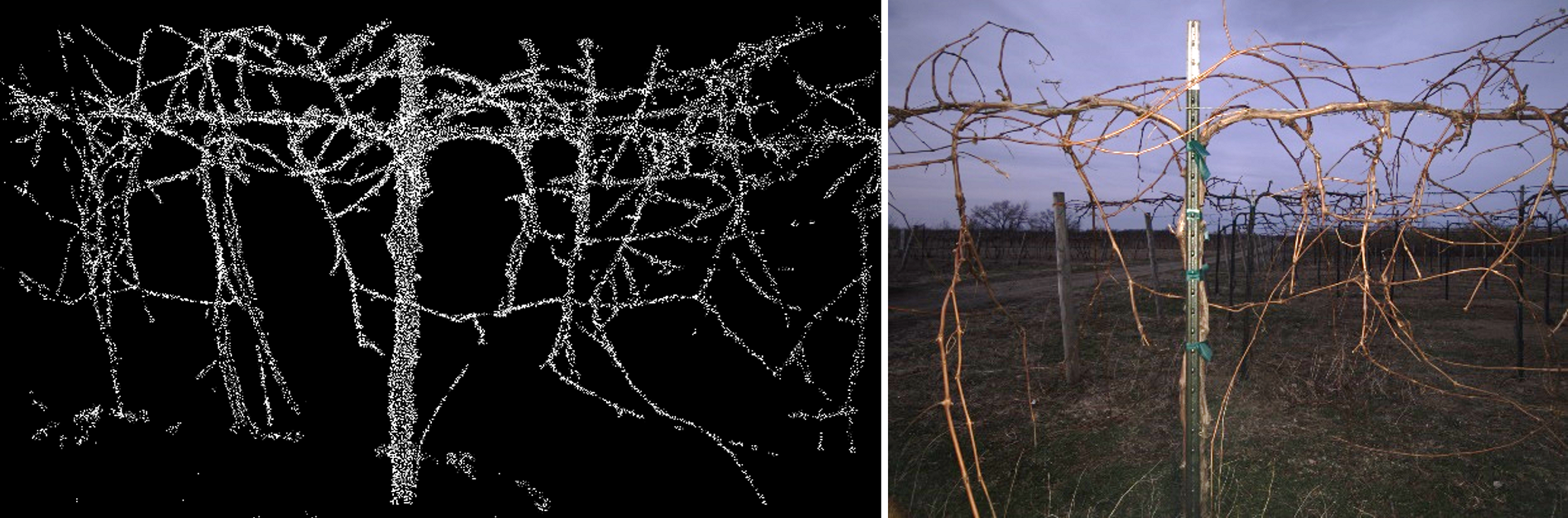}
    %\hspace{.2in}

    \caption{Complex vine registration with multi-camera scan match.}
    \label{fig:complex_vine}
\end{figure}

In this study, complex vines structures were simplified by manually pre-pruning with a machine. This step not only facilitated the perception pipeline, but the overall pruning operation. Despite the heuristic-based choices of multiple parameters, the 3D reconstruction method seems to be applicable to uncut, highly vigorous and cluttered vines (see Fig. \ref{fig:complex_vine}). However, it can be argued that such vine could potentially have higher occlusion that could lead to incomplete or missing canes and affect TPE. This limitation could be handled with an in-hand camera system to explore regions of high occlusion and iteratively add missing links. In recent history, deep learning-based point cloud registration \cite{elbaz20173d} have shown promising results to register noisy point  cloud data without accurate initial alignments and could potentially eliminate frequent calibration and initialization requirements. However, such a supervised approach could potentially require larger training samples to achieve good results.

Large datasets to training machine learning models for deep learning-based computer vision is a bottleneck in specialty crop industry and agriculture, in general. The combination of vast amounts of cultivars and variations within those varieties makes collection and maintenance of labeled datasets for supervised machine learning extremely challenging. The consistency in image exposure and color achieved with the active light camera proved to greatly reduce variance caused in images by ambient lighting. In consequence, the training sample size was reduced by multiple folds to achieve similar bud detection results when compared to models trained with larger datasets (see \cite{silwal2021robust} for more details). The availability of public datasets with 3D plant models with proximal sensors are even rarer in agriculture. As hand-labeling of large 3D dataset to segment canes from rest of the vines was very resource intensive, we refrained from the state-of-the-art deep networks and opted to classical machine learning with SVM. With the combination of singular value decomposition and SVM as a binary classifier, the region growing algorithm was robust at segmenting dormant canes. The training of the SVM model required relatively small but hand-engineered features, and the generation of data for training and the training process itself could be done in a few minutes. Throughout our three years of development and testing, we have only required to train the model once, and it seemed to work equally well on simple as well as complex vine structures. The region growing algorithm essentially exploits the nature of vines. It utilizes buds that are naturally present in vines as seeding points for growing regions in the segmentation process. As all vine varieties have canes with buds, with minor tweaks we expect this algorithm to be adaptable to most vine architectures with relatively small amount of data for retraining. This could be significantly useful as vines are high-value crops and vine industry plays a major role in specialty crops industry. %To facilitate and encourage further research in this area, all data (images and point clouds) collected in this study will be publicly shared.

When pruning a vine, professional workers only keep canes that are healthy and within a certain diameter. These quality attributes of canes are currently not included in our work. Another key limitation in our current computer vision pipeline is the potential effects from wind. For the latter, we observed average wind speed up to 12 miles/hour (mph) (5.3 m/s) and some gusts up to 25 mph (11 m/s). As vines in commercial vineyards are very rigidly supported by metal posts (vertically) and trellis wire(s) horizontally, small winds gusts (up to 12 mph or 5.3 m/s) seemed to have minimum effect in the 3D reconstruction of pre-pruned vines. However, higher wind speeds could arguably cause significant issues in the registration process for any scan match-based approach. Especially in vines with longer canes (not pre-pruned) and regions further away from the rigid trunk and cordon supports. To minimize such affect, we selected spur pruning to retain fewer number of buds per cane where the cut-points were closer to the supportive structures and minimally affected by wind. For other pruning rules such as cane pruning where longer sections of cane need to be retained, affects from wind could be a significant problem. For the former, we currently consider all canes as healthy and viable. Although measuring cane diameter is relatively straight forward from the stereo images, assessing cane quality/ health would require additional sensing ability. More advanced camera systems such as hyper-spectral or thermal imaging technologies in conjunction with end-to-end deep networks could potentially provide robust solutions. 

The addition of a prismatic base to the kinematic chain of the 6-DoF robot arm seems to have added several advantages. The motion planning and execution part with initial and final approach to the pruning locations were attempts to generate natural-looking motion of the arm as well as to cautiously interact with the vine structure. The initial planner was RRT-connect that positioned the end-effector 15 cm from the final destination. It converged to a solution in almost all cases (99\%). However, the shortcomings in the manipulation end in this work are mainly attributed to the Cartesian path planner where 100\% of the interpolated trajectory were sometimes not achievable. Furthermore, in some cases, the ROS’s inbuilt Cartesian planner generated jerky motions that in some cases caused the tip/side of the end-effector to push the cane rather than securing it in-between the cutting blades. %It was difficult to tune the jump-factor parameter in the Cartesian planner that worked in all cases.  

The current state-of-the-art and the interest of the research community to control robot arm in complex environment is reinforcement learning (RL). Similar to Deep learning in computer vision, Deep Reinforcement Learning policies tend to provide end-to-end solutions to manipulation tasks and could likely reduce dependency on heuristically set parameters and behaviors for pruning. Furthermore, with Deep RL more sophisticated capabilities that are not possible with existing sampling-based methods such as learning to prune from expert demonstrations could generalize pruning across different vine varieties and architectures. Our recent efforts in using RL policies to manipulate robot arm for pruning can be found in \cite{yandun2021deeprl}. 

As described in Section \ref{sec::reachability}, to prune the remaining 32\% of vines, the robot had to repeat all processes from the other side. Although, a robot arm with longer reach could solve this issue, vine architectures with uniform cane distribution on a single side and well-defined vine to vine separation would be advantageous. Furthermore, viticultural practices are critical for reliable robotic pruning operations. Cane and spur pruning are the main pruning methods adopted by the industry. However, to maintain balance between yield, quality, and vegetative growth, accurate estimation of vine size is necessary. The estimation of vine size which is often done by pruning weight estimation \cite{balancePruning} determines the number of buds to retain per vine. This method of pruning is formally referred to as balance pruning, as the amount to prune is based on the capacity of the individual vine \cite{balancePruning}. None of the existing robotic prototypes, including this work, have implemented such a strategy. However, our approach of using bud detection and pruning strategy to retaining a fixed number of buds per cane sets us on the right path to achieving balanced pruning.

Finally, we selected MPC controller for navigation in this work mainly for the four following reasons: i) it has produced good results in autonomous navigation for a variety of vehicles and driving conditions \cite{SakhdariMPC,amer2017modelling}, ii) its formulation naturally allows to constraint the optimization problem to obtain desired practical results, iii) it produced a smoother navigation in terms of overshooting and cross-track error when compared with approaches like pure pursuit in prior tests, and iv) as it is a model-driven strategy we can increase its complexity including the dynamics of the vehicle or other variables for future research. It is worth noting that the second point was particularly useful for this application, as we limited the control effort to obtain maneuvers that reduced the risk of damaging the surrounding vegetation. Although, just GPS way point following seemed robust for this application, inclusion of local sensor for navigation as well as safety critical features such as obstacle detection \& avoidance, and compliance to farm vehicle and field workers are currently ongoing developments. 

\section{Conclusions}  
%Inconsistency in pruning becomes a source of imbalance in vines that later affects fruit quality, yield and other canopy management tasks throughout the growing season. Furthermore, our economic study \ref{sec::economic_study} shows the potential of nearly eight times cost reduction in pruning with autonomous robots. We then conclude that a selective pruning robot is a clear need in the grapevine industry to improve productivity. 

In this work we presented a combination of tools, techniques, and system development details of an autonomous vine pruning robot as a follow-up pruner. Highly vigorous Concord vines in a commercial vineyard were mechanically pre-pruned to ease robotic operations. The foci of attention here were not only to develop a system mostly utilizing off-the-shelf hardware components for a proof-of-concept prototype, but mostly to understand what is takes to robotically prune grape vines. The key technical challenges that we addressed in this work were robust imaging capability in the outdoors and data efficient machine learning models for processing vine structures. The illumination invariant camera system proved to be a valuable component as consistent image data were acquired at any lighting condition. This also led to fewer training sample for detecting buds in images and eased 3D reconstruction. Results from the field study show that even complex vines structures could be accurately modeled from single side imaging. The integrated system robustly identified pruning location and  pruned 87\% of the canes successfully, with an average cycle time of 213 sec /vine from two sides and 137 sec/vine from one side. Improved pruning efficiency will require robustness in manipulation and advanced sensing capabilities to assess cane health and vine size for balance pruning. The mechanical design with redundant manipulator was enough to address a single vine and could have multiple uses throughout the growing season, such as selective shoot thinning and harvesting.

%============================================================
\subsubsection*{Acknowledgments}
This research was supported by United States Department of Agriculture, National Institute of Food and Agriculture, Specialty Crop Research Initiative under project no: USDA-NIFA-SCRI-2015-09334. The authors would like to acknowledge Zania Pothen, Tanvir Parhar, Harjatin Baweja and the CLEREL field staff for their coordination and support during the field experiments.

\bibliographystyle{apalike}
\bibliography{fr_abhi_refs}
\clearpage
%============================================================
\subsubsection*{Appendix: A brief economic study to compare cost between different pruning technologies }
\label{sec::economic_study}
This section summarizes a brief economic study that highlights the cost associated with manual pruning operation and the benefits of robotic pruning both as a hand follow-up operation and fully independent and autonomous system. Table \ref{tab:vineyard_measurments} shows typical vineyard dimensions for three varieties of vine including Riesling, Concord, and Vignoles that are commonly grown across the United States. The variables evaluated include row width, vine separation, cordon height, number of trellis wire and average number of vines per acre. These measurements were taken at the same commercial vineyard site used for testing the robotic pruner described in this paper. To quantify human pruning speed, an experienced professional pruner was asked to prune 30 sample of each vine variety at a regular pace. On average, it took 150 seconds to prune a vine with an average of 14 canes to cut.

\begin{table}[!h]
\centering
\caption{\label{tab:vineyard_measurments} Various vine architecture measurements in the test vineyards.}
\vspace{10pt}
\begin{tabular}{ccccc}
\rowcolor[HTML]{C0C0C0} 
\cellcolor[HTML]{C0C0C0} &
  \multicolumn{3}{c}{\cellcolor[HTML]{C0C0C0}Vine type} &
  \cellcolor[HTML]{C0C0C0} \\
\rowcolor[HTML]{C0C0C0} 
\multirow{-2}{*}{\cellcolor[HTML]{C0C0C0}Measurement   unit: feet (meter)} &
  Riesling &
  Vignoles &
  Concord &
  \multirow{-2}{*}{\cellcolor[HTML]{C0C0C0}Average} \\
\rowcolor[HTML]{EFEFEF} 
Row width (W)                           & 9 (2.74) & 9 (2.74) & 9 (2.74) & 9 (2.74) \\
Vine separation (L)                     & 6 (1.83) & 4 (1.22) & 9 (2.74) & 6 (1.83) \\
\rowcolor[HTML]{EFEFEF} 
Cordon height (H)                       & 2 (0.61) & 6 (1.83) & 6 (1.83) & 6 (1.83) \\
No. of trellis                          & 4        & 1        & 1        & -        \\
\rowcolor[HTML]{EFEFEF} 
No of vines per Acre                    & 807      & 1210     & 537      & 850      \\
Average cane per vine                   & 11       & 13       & 18       & 14       \\
\rowcolor[HTML]{EFEFEF} 
Average manual pruning speed (sec/vine) & 142.8    & 95.4     & 214.2    & 150     
\end{tabular}
\end{table}

\begin{figure} [!h]
    \centering
    \includegraphics[height=2.25in]{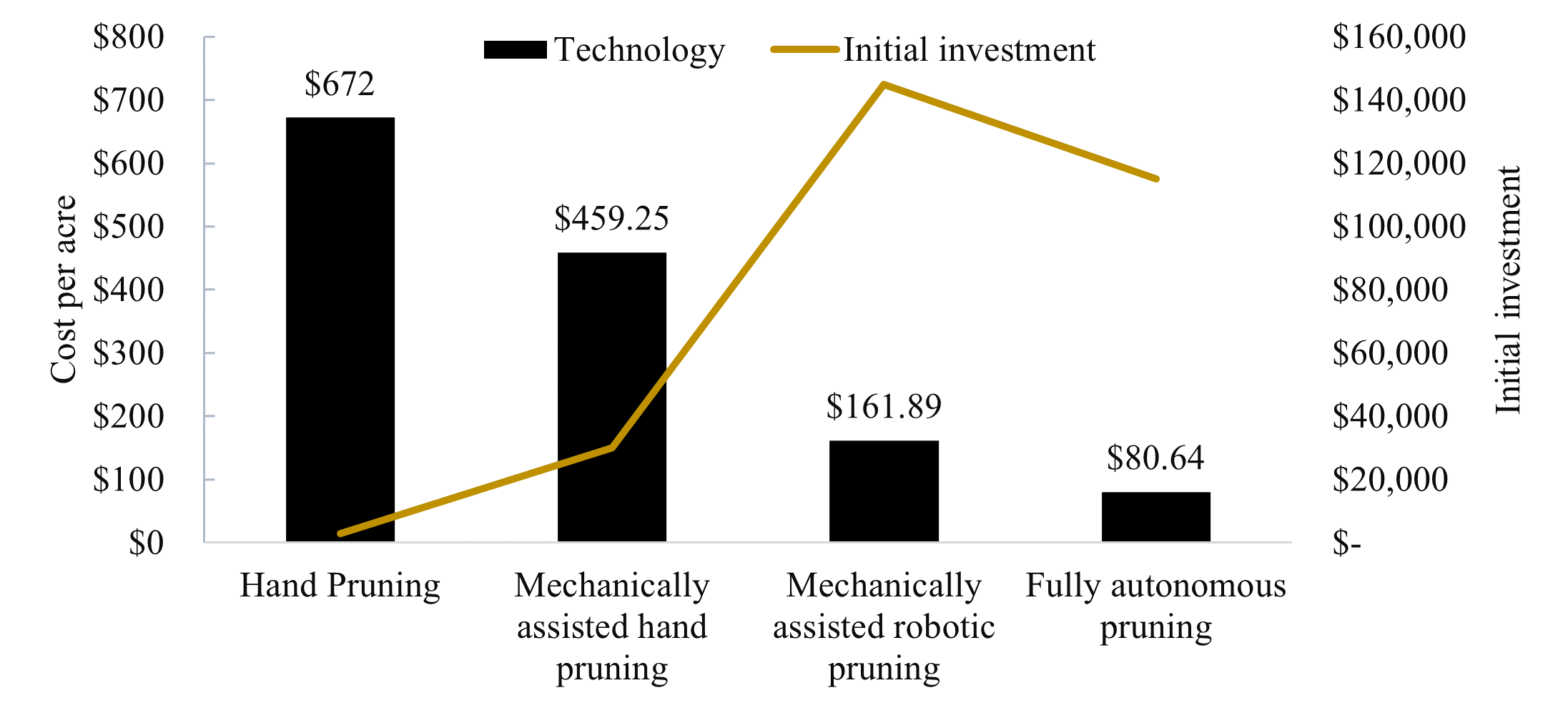}
    \hspace{.2in}

    \caption{Cost breakdown for different pruning technologies.}
    \label{fig:cost_breakdown}
\end{figure}

The numeric values (see Table \ref{tab:cost_hand_pruning} \& \ref{tab:robotic_pruning})used in this brief study are mostly inspired from \cite{economicStudy}. Table \ref{tab:cost_hand_pruning} and \ref{tab:robotic_pruning} have more details on additional numeric values and their function in the calculation and comparison of four types of pruning technology viz. hand pruning, mechanically assisted hand pruning, mechanically assisted robotic pruning, and fully autonomous pruning. Additionally, we also assumed 25\% increase on any outdated costs and wages used by \cite{economicStudy}. We also assume minimum wages of \$15 per hour (based on NY minimum wage  \cite{newyorkstate2021})) and cost of fuel is taken as national average of \$2.38 for 2021 from \cite{NationalAverage}. For robotic pruning, we assume the robot runs on fuel for 18 hours long operation to match the labor counterparts of three pruning crew working 6 hours each. The calculations are detailed in Table \ref{tab:cost_hand_pruning} and \ref{tab:robotic_pruning}. Our calculations show that the cost of hand pruning per acre was \$672, which was the highest cost amongst all technologies, as expected. In mechanically assisted pruning, a dedicated hardware system pre-prunes vines that ease the work environment for pruners at the cost of the mechanical system. Here, the cost of mechanically assisted and hand follow-up operation together summed up to\$459.25 per acre. In mechanically assisted robotic pruning, robots autonomously perform the follow-up pruning operation after vines are mechanically pre-pruned. Although at a glance it seems to increase initial investments, it drastically decreased per acre cost of pruning to \$161.9. With a fully autonomous system capable of pruning vines end-to-end without any assistance, the cost of pruning goes down to just \$80.64 per acre. As evident from Table \ref{tab:cost_hand_pruning} and \ref{tab:robotic_pruning}, and depicted in Fig. \ref{fig:cost_breakdown}, assisted robotic technology or fully independent robotic solutions could decrease cost of pruning by nearly 4 to 8 times respectively. Thus, robotic systems for pruning has clear advantage to existing technology/ practice and needs in today’s economy and in the long-term could prove to be  profitable.

%=======================================================
\begin{table}[!h]
\centering
\caption{\label{tab:cost_hand_pruning} Cost associated with hand and mechanically assisted pruning.}
\vspace{10pt}
%\resizebox{\textwidth}{!}{%
\begin{tabular}{cccc}
\rowcolor[HTML]{C0C0C0} 
\multicolumn{2}{c}{\cellcolor[HTML]{C0C0C0}} &
  \multicolumn{2}{c}{\cellcolor[HTML]{C0C0C0}Mechanically assisted pruning} \\
\rowcolor[HTML]{C0C0C0} 
\multicolumn{2}{c}{\multirow{-2}{*}{\cellcolor[HTML]{C0C0C0}Hand pruning}} &
  Mechanical pre-pruning &
  Hand follow-up \\
\cellcolor[HTML]{EFEFEF}Labor (Hours/ Acre) &
  \cellcolor[HTML]{EFEFEF}32 hrs. &
   &
  \cellcolor[HTML]{EFEFEF} \\
Minimum wage (per hour) &
  \$15 &
   &
  \cellcolor[HTML]{EFEFEF} \\
\cellcolor[HTML]{EFEFEF}Benefit &
  \cellcolor[HTML]{EFEFEF}40\% &
   &
  \cellcolor[HTML]{EFEFEF} \\
Direct cost (per hour) & \$21 & \multirow{-4}{*}{0.65 equipment hrs. + 2.4 labor hour} & \multirow{-4}{*}{\cellcolor[HTML]{EFEFEF}3 laborers for total of 18 hrs.} \\
\cellcolor[HTML]{EFEFEF}Total Cost (per acre) &
  \cellcolor[HTML]{EFEFEF}\textbf{\$672} &
  \textbf{\$81.25} &
  \textbf{\$378} \\
\multicolumn{2}{c}{} &
   &
   \\
\multicolumn{2}{c}{\multirow{-2}{*}{}} &
  \multirow{-2}{*}{Combined Cost   (per acre)} &
  \multirow{-2}{*}{\textbf{\$459.25}} \\
\rowcolor[HTML]{EFEFEF} 
Initial investment* &
  - &
  \$30,000 &
  -
\end{tabular}%
%}
\end{table}
%=========================================================================
\begin{table}[!h]
\centering
\caption{\label{tab:robotic_pruning}Cost associated with robotic pruning.}
\vspace{10pt}
%\resizebox{\textwidth}{!}{%
\begin{tabular}{ccc}
\rowcolor[HTML]{C0C0C0} 
Fully   Autonomous Pruning                  &                           & Mechanical   Pre-pruning \\
\cellcolor[HTML]{EFEFEF}Labor (Hours/ Acre) & \cellcolor[HTML]{EFEFEF}- &                          \\
Minimum wage (per hr.)                     & -                         &                          \\
\cellcolor[HTML]{EFEFEF}Benefit             & \cellcolor[HTML]{EFEFEF}- &                          \\
Run time                                    & 18 hrs.                   &                          \\
\cellcolor[HTML]{EFEFEF}Robot fuel (9 gals @ 0.5   gals/ hr.) & \cellcolor[HTML]{EFEFEF}\$21.42 &                                                        \\
Generator fuel (9 gals @ 0.5   gals/ hr.)   & \$21.42                   &                          \\
\cellcolor[HTML]{EFEFEF}Lubrication cost (\$2.1/ hr.)         & \cellcolor[HTML]{EFEFEF}\$37.80 & \multirow{-7}{*}{0.65 equipment hrs. + 2.4 labor hour} \\
Total Cost (per acre)                       & \textbf{\$80.64}          & \$81.25                  \\
\rowcolor[HTML]{EFEFEF} 
Combined Cost (per acre)                    &                           & \textbf{\$161.90}        \\
Initial investment*                         & \textbf{\$115,000}        & \$30,000                
\end{tabular}%
%}

\begin{tablenotes}
  \item *The robotic pruning prototype is estimated to cost \$115,000 US. The \$30,000 US cost is the estimated value of the OXBO pruning head attachable to farm vehicles.
\end{tablenotes}
\end{table}

%    \begin{tablenotes}
%      \item *The robotic pruning prototype is estimated to cost \$115,000 US. The \$30,000 US cost is the estimated value of the OXBO pruning head attachable to farm vehicles.
%    \end{tablenotes}
    
\end{document}